\documentclass[10pt,twocolumn,letterpaper]{article}

\usepackage{cvpr}
\usepackage{times}
\usepackage{epsfig}
\usepackage{graphicx}
\usepackage{amsmath}
\usepackage{amssymb}

\usepackage{mathtools}
\usepackage{acronym}
\usepackage{caption}
\usepackage{array}
\usepackage{tabularx}
\usepackage{subcaption}
\usepackage{booktabs}
\usepackage{multicol}
\usepackage{multirow}

\acrodef{SPADE}{Spatially-Adaptive Normalization}
\acrodef{AdaIN}{Adaptive Instance Normalization}
\acrodef{LRN}{Local Response Normalization}
\acrodef{BN}{Batch Normalization}
\acrodef{IN}{Instance Normalization}
\acrodef{LN}{Layer Normalization}
\acrodef{GN}{Group Normalization}
\acrodef{WN}{Weight Normalization}
\acrodef{Conditional BN}{Conditional Batch Normalization}

\acrodef{mIoU}{mean Intersection-over-Union}
\acrodef{accu}{pixel accuracy}
\acrodef{FID}{Fréchet Inception Distance}

\def\facade{fa\c{c}ade\xspace}

\def\Facades{Fa\c{c}ades\xspace}

\usepackage[pagebackref=true,breaklinks=true,letterpaper=true,colorlinks,bookmarks=false]{hyperref}

\cvprfinalcopy 

\def\cvprPaperID{} 

\begin{document}

\title{SEAN: Image Synthesis with Semantic Region-Adaptive Normalization}

\author{Peihao Zhu\textsuperscript{1} \quad Rameen Abdal\textsuperscript{1} \quad Yipeng Qin\textsuperscript{2} \quad Peter Wonka\textsuperscript{1} \\
\\
\textsuperscript{1}KAUST \quad \textsuperscript{2}Cardiff University
}


\twocolumn[{%
\renewcommand\twocolumn[1][]{#1}%
\maketitle
\thispagestyle{empty}

\begin{center}
    \includegraphics[width=\linewidth]{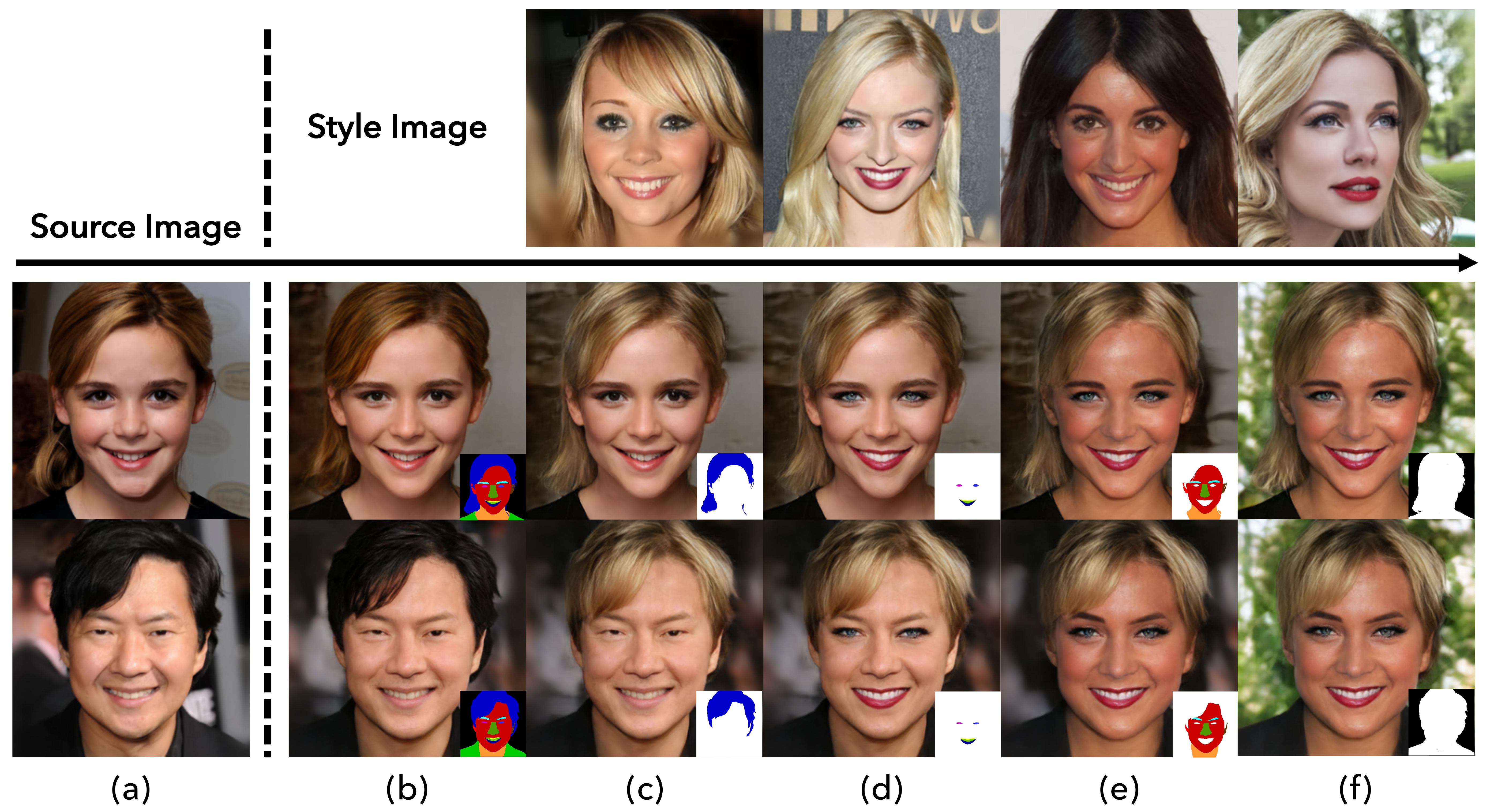}
    \vspace*{-5mm}
    \captionof{figure}{Face image editing controlled via style images and segmentation masks. a) source images. b) reconstruction of the source image; segmentation mask shown as small inset. c - f) four separate edits; we show the image that provides new style information on top and show the part of the segmentation mask that gets edited as small inset. The results of the successive edits are shown in row two and three. The four edits change hair, mouth and eyes, skin tone, and background, respectively.}
    \label{fig:teaser}
\end{center}%
}]

\begin{abstract}
    We propose semantic region-adaptive normalization (SEAN), a simple but effective building block for Generative Adversarial Networks conditioned on segmentation masks that describe the semantic regions in the desired output image.
    Using SEAN normalization, we can build a network architecture that can control the style of each semantic region individually, e.g., we can specify one style reference image per region. SEAN is better suited to encode, transfer, and synthesize style than the best previous method in terms of reconstruction quality, variability, and visual quality.
    We evaluate SEAN on multiple datasets and report better quantitative metrics (e.g. FID, PSNR) than the current state of the art.
    SEAN also pushes the frontier of interactive image editing. We can interactively edit images by changing segmentation masks or the style for any given region. We can also interpolate styles from two reference images per region. Code:  \small{\url{https://github.com/ZPdesu/SEAN}}.
\end{abstract}

\begin{figure*}[th]
\centering
\includegraphics[width=\linewidth]{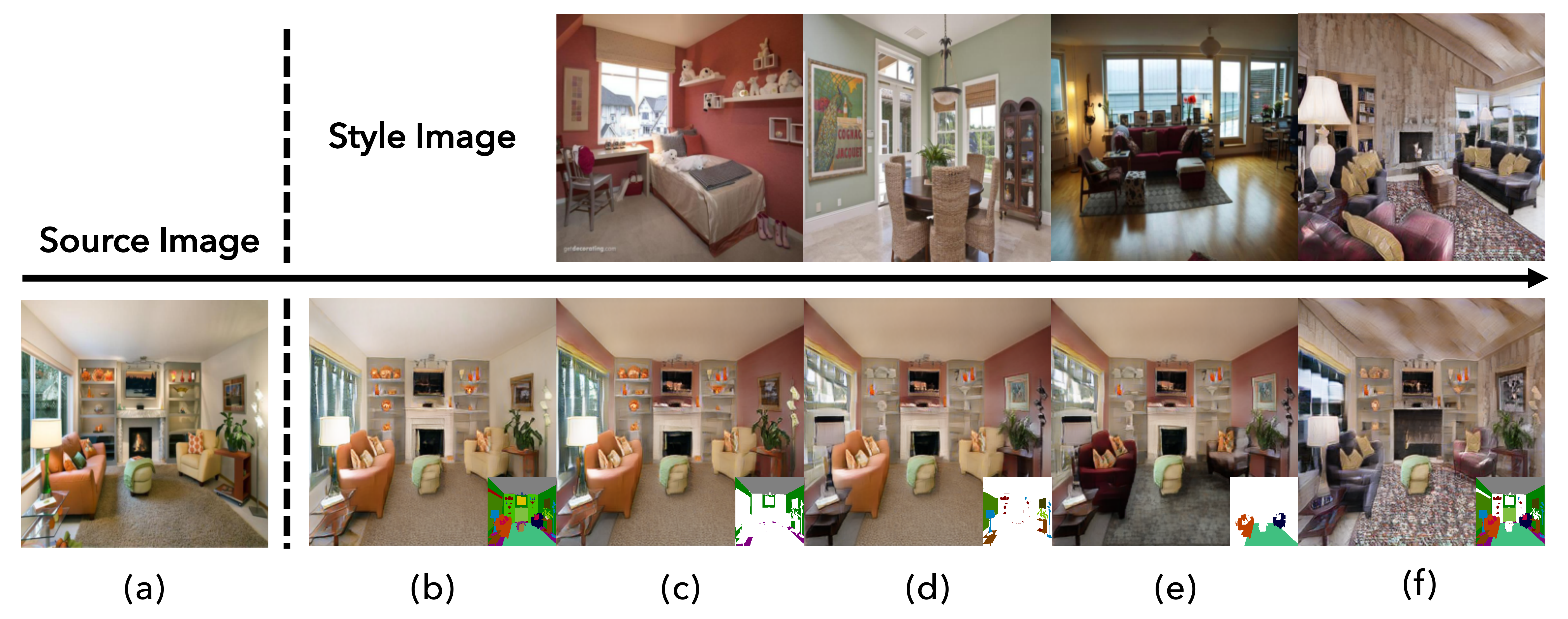}
\vspace*{-7mm}
\caption{Editing sequence on the ADE20K dataset. (a) source image, (b) reconstruction of the source image, (c-f) various edits using style images shown in the top row. The regions affected by the edits are shown as small insets.}
\label{fig:ADE editing}
\end{figure*}

\section{Introduction}

In this paper we tackle the problem of synthetic image generation using conditional generative adversarial networks (cGANs). Specifically, we would like to control the layout of the generated image using a segmentation mask that has labels for each semantic region and ``add'' realistic styles to each region according to their labels . For example, a face generation application would use region labels like eyes, hair, nose, mouth, \etc~and a landscape painting application would use labels like water, forest, sky, clouds, \etc
While multiple very good frameworks exist to tackle this problem~\cite{isola2016imagetoimage,chen2017photographic,qi2018semiparametric, wang2018pix2pixHD}, the currently best architecture is SPADE~\cite{park2019SPADE} (also called GauGAN). Therefore, we decided to use SPADE as starting point for our research. By analyzing the SPADE results, we found two shortcomings that we would like to improve upon in our work.

First, SPADE uses only one style code to control the entire style of an image, which is not sufficient for high quality synthesis or detailed control.
For example, it is easily possible that the segmentation mask of the desired output image contains a labeled region that is not present in the segmentation mask of the input style image. 
In this case, the style of the missing region is undefined, which yields low quality results.
Further, SPADE does not allow using a different style input image for each region in the segmentation mask. 
Our first main idea is therefore to control the style of each region individually, \ie, our proposed architecture accepts one style image per region (or per region instance) as input.

Second, we believe that inserting style information only in the beginning of a network is not a good architecture choice. Recent architectures~\cite{karras2018stylebased,liu2019fewshot,alharbi2019latent} have demonstrated that higher quality results can be obtained if style information is injected as normalization parameters in multiple layers in the network, e.g. using AdaIN~\cite{huang2017arbitrary}. However, none of these previous networks use style information to generate spatially varying normalization parameters.
To alleviate this shortcoming, our second main idea is to design a normalization building block, called SEAN, that can use style input images to create spatially varying normalization parameters per semantic region. An important aspect of this work is that the spatially varying normalization parameters are dependent on the segmentation mask as well as the style input images.

Empirically, we provide an extensive evaluation of our method on several challenging datasets: CelebAMask-HQ~\cite{CelebAMask-HQ,karras2017progressive,liu2015faceattributes}, 
CityScapes~\cite{Cordts2016Cityscapes}, ADE20K~\cite{zhou2017scene}, and our own \Facades dataset. 
Quantitatively, we evaluate our work on a wide range of metrics including FID, PSNR, RMSE and segmentation performance; qualitatively, we show examples of synthesized images that can be evaluated by visual inspection. 
Our experimental results demonstrate a large improvement over the current state-of-the-art methods. In summary, we introduce a new architecture building block \emph{SEAN} that has the following advantages:
\begin{enumerate}
  \item SEAN improves the quality of the synthesized images for conditional GANs. We compared to the state of the art methods SPADE and Pix2PixHD and achieve clear improvements in quantitative metrics (e.g. FID score) and visual inspection.
  \item SEAN improves the per-region style encoding, so that reconstructed images can be made more similar to the input style images as measured by PSNR and visual inspection.
  \item SEAN allows the user to select a different style input image for each semantic region. This enables image editing capabilities producing much higher quality and providing better control than the current state-of-the-art methods. Example image editing capabilities are interactive region by region style transfer and per-region style interpolation (See Figs.~\ref{fig:teaser}, \ref{fig:ADE editing}, and \ref{fig:style interpolation}).
\end{enumerate}

\section{Related Work}

\vspace*{2mm}\noindent {\bf Generative Adversarial Networks.}
Since their introduction in 2014, Generative Adversarial Networks (GANs)~\cite{goodfellow2014generative} have been successfully applied to various image synthesis tasks, \eg image inpainting~\cite{yu2018generative,demir2018patch}, image manipulation \cite{zhu2016generative,Bau:Ganpaint:2019,abdal2019image2stylegan} and texture synthesis~\cite{TextureSynthesis2016,slossberg2018high,Fr_hst_ck_2019}.
With continuous improvements on GAN architecture~\cite{radford2015unsupervised,karras2018stylebased,park2019SPADE}, loss function~\cite{Mao_2017,arjovsky2017wasserstein} and regularization~\cite{gulrajani2017improved,miyato2018spectral,mescheder2018training}, the images synthesized by GANs are becoming more and more realistic.
For example, the human face images generated by StyleGAN~\cite{karras2018stylebased} are of very high quality and are almost indistinguishable from photographs by untrained viewers.
A traditional GAN uses noise vectors as the input and thus provides little user control. This motivates the development of conditional GANs (cGANs)~\cite{mirza2014conditional} where users can control the synthesis by feeding the generator with conditioning information.
Examples include class labels \cite{miyato2018cgans,mescheder2018training,brock2018large}, text~\cite{reed2016generative,hong2018inferring,xu2018attngan} and images~\cite{isola2016imagetoimage,zhu2017unpaired,liu2017unsupervised,wang2018pix2pixHD,wang2018pix2pixHD,park2019SPADE}.
Our work is built on the conditional GANs with image inputs, which aims to tackle image-to-image translation problems.

\vspace*{2mm}\noindent {\bf Image-to-Image Translation.}
Image-to-image translation is an umbrella concept that can be used to describe many problems in computer vision and computer graphics.
As a milestone, Isola \etal~\cite{isola2016imagetoimage} first showed that image-conditional GANs can be used as a general solution to various image-to-image translation problems.
Since then, their method is extended by several following works to scenarios including: unsupervised learning ~\cite{zhu2017unpaired,liu2017unsupervised}, few-shot learning ~\cite{liu2019fewshot}, high resolution image synthesis~\cite{wang2018pix2pixHD}, multi-modal image synthesis~\cite{zhu2017toward,huang2018multimodal} and multi-domain image synthesis~\cite{choi2017stargan}.
Among various image-to-image translation problems, semantic image synthesis is a particularly useful genre as it enables easy user control by modifying the input semantic layout image~\cite{CelebAMask-HQ,Bau:Ganpaint:2019,gu2019maskguide,park2019SPADE}.
To date, the SPADE~\cite{park2019SPADE} model (also called GauGAN) generates the highest quality results. In this paper, we will improve SPADE by introducing per-region style encoding.

\vspace*{2mm}\noindent {\bf Style Encoding.}
Style control is a vital component for various image synthesis and manipulation applications~\cite{GatysStyle2015,TextureSynthesis2016,Adain2017,karras2018stylebased,abdal2019image2stylegan}.
\textit{Style} is generally not manually designed by a user, but extracted from reference images.
In most existing methods, styles are encoded in three places: i) statistics of image features~\cite{GatysStyle2015,sendik2017deep}; 
ii) neural network weights (\eg fast style transfer~\cite{johnson2016perceptual,zhou2018non,yang2019controllable});
iii) parameters of a network normalization layer~\cite{Adain2017,Kotovenko2019disentangleStyle}).
When applied to style control, the first encoding method is usually time-consuming as it requires a slow optimization process to match the statistics of image features extracted by image classification networks ~\cite{GatysStyle2015}. 
The second approach runs in real-time but each neural network only encodes the style of selected reference images~\cite{johnson2016perceptual}.
Thus, a separate neural network is required to be trained for each style image, which limits its application in practice.
To date, the third approach is the best as it enables arbitrary style transfer in real-time~\cite{Adain2017} and it is used by high-quality networks such as StyleGAN~\cite{karras2018stylebased}, FUNIT~\cite{liu2019fewshot}, and SPADE~\cite{park2019SPADE}.
Our per-region style encoding also builds on this approach. We will show that our method generates higher quality results and enables more-detailed user control.

\section{Per Region Style Encoding and Control}

Given an input style image and its corresponding segmentation mask, this section shows i) how to distill the per-region styles from the image according to the mask and ii) how to use the distilled per-region style codes to synthesize photo-realistic images.

\subsection{How to Encode Style?}

\vspace*{2mm}\noindent {\bf Per-Region Style Encoder.} To extract per region styles, we propose a novel style encoder network to distill the corresponding style code from each semantic region of an input image simultaneously (See the subnetwork \emph{Style Encoder} in Figure~\ref{fig:generator} (A)).
The output of the style encoder is a $512 \times s$ dimensional style matrix $\mathbf{ST}$, where $s$ is the number of semantic regions in the input image. Each column of the matrix corresponds to the style code of a semantic region.
Unlike the standard encoder built on a simple downscaling convolutional neural network, our per-region style encoder employs a ``bottleneck'' structure to remove the information irrelevant to styles from the input image.
Incorporating the prior knowledge that styles should be independent of the shapes of semantic regions, we pass the intermediate feature maps ($512$ channels) generated by the network block \emph{TConv-Layers} through a region-wise average pooling layer and reduce them to a collection of $512$ dimensional vectors.
As implementation detail we would like to remark that we use $s$ as the number of semantic labels in the data set and set columns corresponding to regions that do not exist in an input image to $0$. As a variation of this architecture, we can also extract style per region instance for datasets that have instance labels, e.g., CityScapes.

\begin{figure}[t]
\centering
\includegraphics[width=\linewidth]{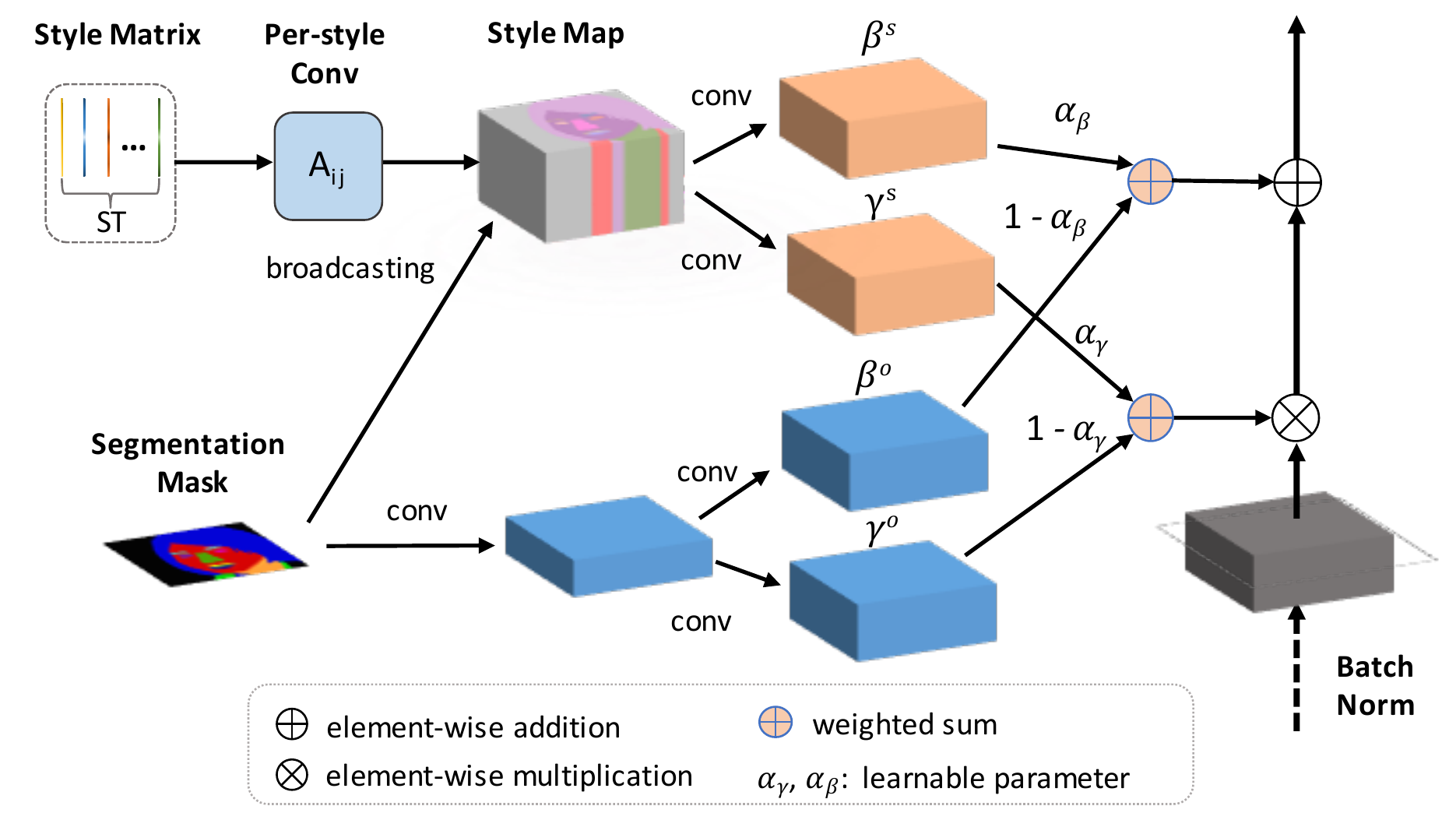}
\caption{SEAN normalization. The input are style matrix $\mathbf{ST}$ and segmentation mask $\mathbf{M}$. In the upper part, the style codes in $\mathbf{ST}$ undergo a per style convolution and are then broadcast to their corresponding regions according to $\mathbf{M}$ to yield a style map. The style map is processed by conv layers to produce per pixel normalization values $\gamma^s$ and $\beta^s$. The lower part (light blue layers) creates per pixel normalization values using only the region information similar to SPADE.}
\label{fig:SEAN normalization}
\end{figure}

\begin{figure*}[t]
\centering
\includegraphics[width=\linewidth]{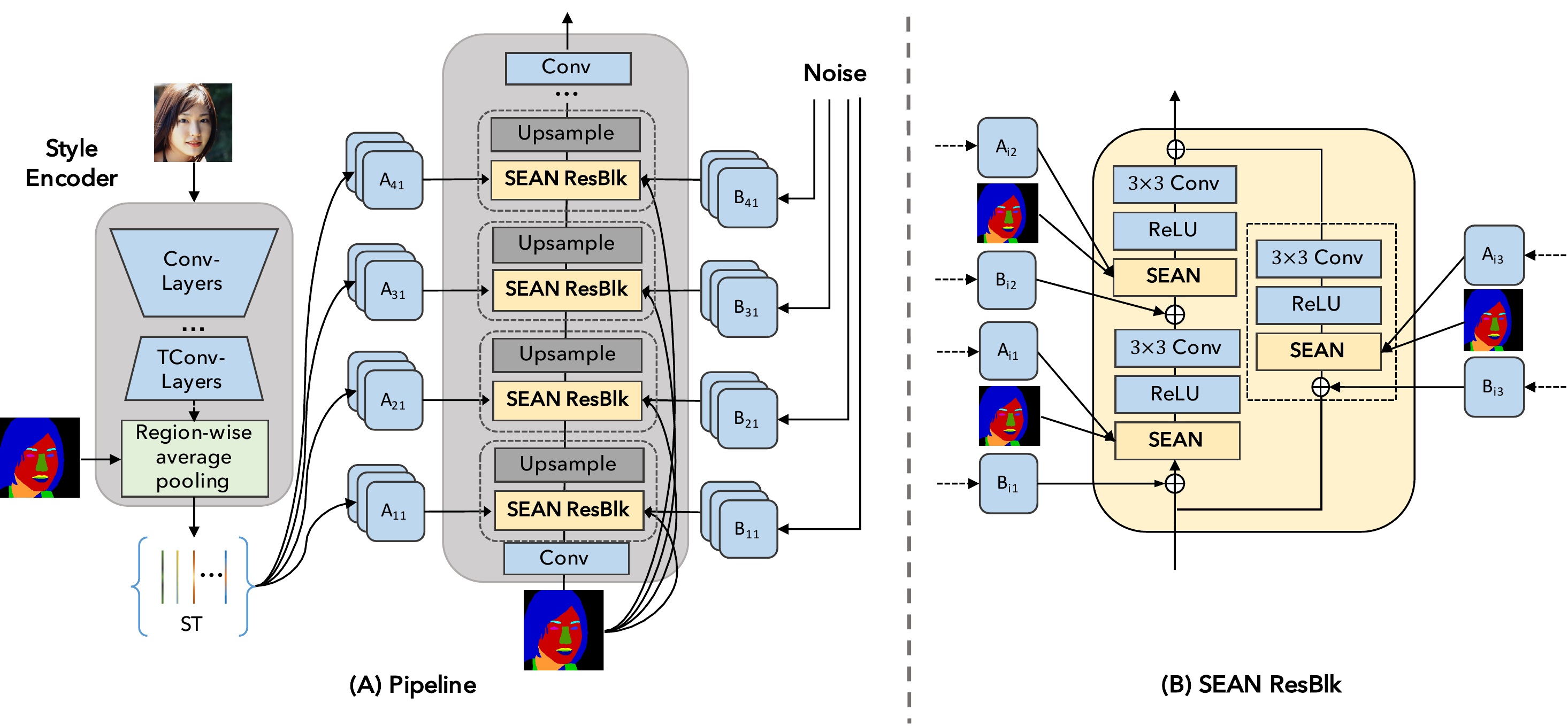}
\caption{SEAN generator. (A) On the left, the style encoder takes an input image and outputs a style matrix $\mathbf{ST}$. The generator on the right consists of interleaved SEAN ResBlocks and Upsampling layers. (B) A detailed view of a SEAN ResBlock used in (A).}
\label{fig:generator}
\end{figure*}

\subsection{How to Control Style?}

With per-region style codes and a segmentation mask as inputs, we propose a new conditional normalization technique called Semantic Region-Adaptive Normalization (SEAN) to enable detailed control of styles for photo-realistic image synthesis.
Similar to existing normalization techniques~\cite{Adain2017,park2019SPADE}, SEAN works by modulating the scales and biases of generator activations.
In contrast to all existing methods, the modulation parameters learnt by SEAN are dependent on both the style codes and segmentation masks.
In a SEAN block (Figure~\ref{fig:SEAN normalization}), a style map is first generated by broadcasting the style codes to their corresponding semantic regions according to the input segmentation mask. 
This style map along with the input segmentation mask are then passed through two separate convolutional neural networks to learn two sets of modulation parameters.
The weighted sums of them are used as the final SEAN parameters to modulate the scales and biases of generator activations.
The weight parameter is also learnable during training.
The formal definition of SEAN is introduced as follows.

\vspace*{2mm}\noindent {\bf Semantic Region-Adaptive Normalization (SEAN).}
A SEAN block has two inputs: a style matrix $\mathbf{ST}$ encoding per-region style codes and a segmentation mask $\mathbf{M}$. 
Let $\mathbf{h}$ denote the input activation of the current SEAN block in a deep convolutional network for a batch of $N$ samples. Let $H$, $W$ and $C$ be the height, width and the number of channels in the activation map.
The modulated activation value at site $\left(n \in N, c \in C, y \in H, x \in W\right)$ is given by

\begin{equation}
\gamma_{c, y, x}(\mathbf{ST,M}) \frac{h_{n, c, y, x}-\mu_{c}}{\sigma_{c}}+\beta_{c, y, x}(\mathbf{ST,M})
\label{eq:activation value}
\end{equation}
where $h_{n, c, y, x}$ is the activation at the site before normalization, the modulation parameters $\gamma_{c, y, x}$ and $\beta_{c, y, x}$ are weighted sums of $\gamma_{c, y, x}^{s}$, $\gamma_{c, y, x}^{o}$ and $\beta_{c, y, x}^{s}$, $\beta_{c, y, x}^{o}$ respectively (See Fig.~\ref{fig:SEAN normalization} for definition of $\gamma$ and $\beta$ variables):
\begin {equation}
\begin{split}
\gamma_{c, y, x}(\mathbf{ST,M}) &= \alpha_{\gamma}\gamma_{c, y, x}^{s}(\mathbf{ST}) + (1-\alpha_{\gamma}) \gamma_{c, y, x}^{o}(\mathbf{M})\\
\beta_{c, y, x}(\mathbf{ST,M}) &= \alpha_{\beta}\beta_{c, y, x}^{s}(\mathbf{ST}) + (1-\alpha_{\beta}) \beta_{c, y, x}^{o}(\mathbf{M})
\label{eq: gamma beta calculation}
\end{split}
\end{equation}
$\mu_{c}$ and $\sigma_{c}$ are the mean and standard deviation of the activation in channel $c$:
\begin{equation}
\mu_{c}=\frac{1}{N H W} \sum_{n, y, x} h_{n, c, y, x}
\label{eq:mu equation}
\end{equation}
\begin{equation}
\sigma_{c}=\sqrt{ \frac{1}{N H W} \left(\sum_{n, y, x}h_{n, c, y, x}^{2}\right) -\mu_{c}^{2}}
\label{eq:sigma equation}
\end{equation}

\begin{table*}[thpb]
\centering
\small
\begin{tabular}{lccc|ccc|ccc|ccc} \toprule

\multirow{2}{*}{Method}   & \multicolumn{3}{c|}{CelebAMask-HQ}  & \multicolumn{3}{c|}{CityScapes}  & \multicolumn{3}{c|}{ADE20K} & \multicolumn{3}{c}{\Facades}\\ \cmidrule{2-13}
{}     & SSIM  & RMSE  & PSNR           & SSIM  & RMSE  & PSNR      & SSIM  & RMSE  & PSNR      & SSIM  & RMSE  & PSNR    \\ \midrule
                       
Pix2PixHD~\cite{wang2018pix2pixHD}  &0.68    &0.15   &17.14         &0.69    &{\bf 0.13}   &18.32       &0.51   &0.22  &13.81       &0.53   &0.16   &16.30 \\
SPADE~\cite{park2019SPADE}  &0.63   &0.21   &14.30         &0.64    &0.18   &15.77       &0.45   &0.28  &11.52       &0.44   &0.22   &13.87 \\
{\bf Ours} &{\bf 0.73}&{\bf 0.12}&{\bf 18.74}  &{\bf 0.70}& {\bf 0.13}& {\bf 18.61} &{\bf 0.58}&{\bf 0.17}&{\bf 16.16} &{\bf 0.58} &{\bf 0.14}& {\bf17.14} \\
\bottomrule

\end{tabular}

\caption{Quantitative comparison of reconstruction quality. Our method outperforms current leading methods using similarity metrics SSIM, RMSE, and PSNR on all the datasets. For SSIM and PSNR, higher is better. For RMSE, lower is better.} 
\label{tab:similarity quantitative table}
\end{table*}

\begin{figure*}[th]
\centering
\includegraphics[width=\linewidth]{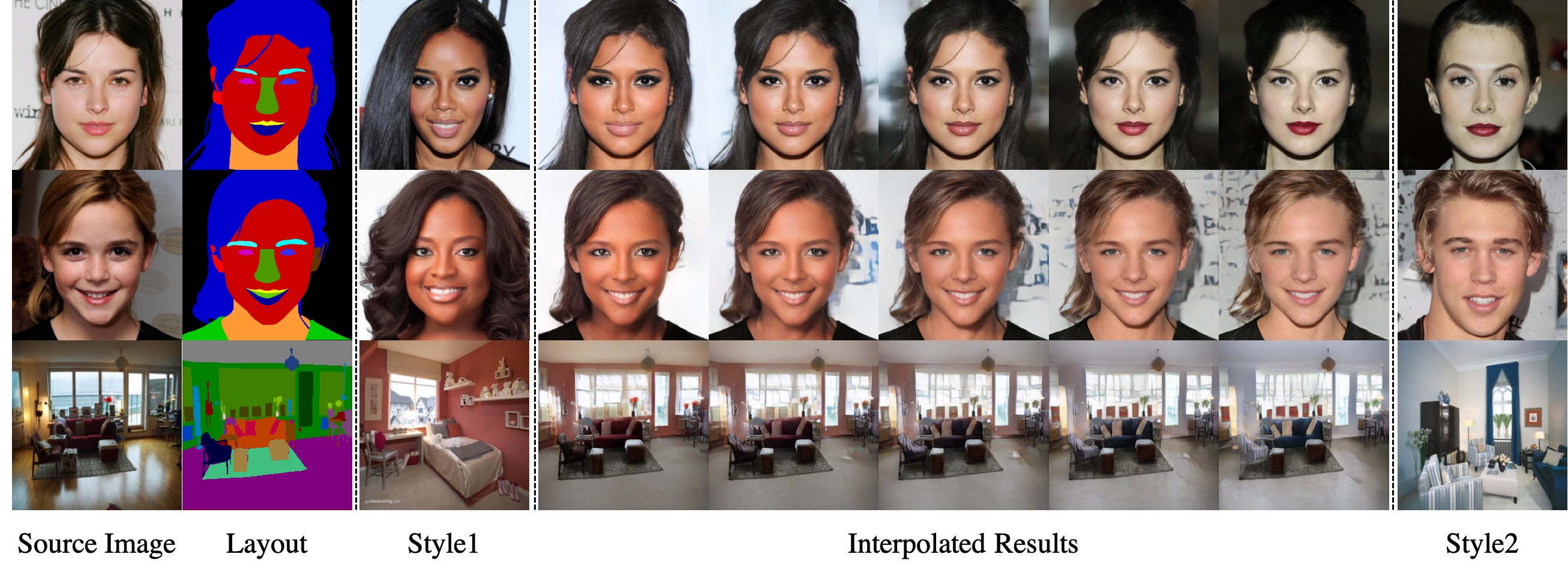}
\vspace*{-7mm}
\caption{Style interpolation. We take a mask from a source image and reconstruct with two different style images (Style1 and Style2) that are very different from the source image. We then show interpolated results of the per-region style codes.}
\label{fig:style interpolation}
\end{figure*}

\begin{table*}[thpb]
\centering
\small
\begin{tabular}{lccc|ccc|ccc|c} \toprule

\multirow{2}{*}{Method}   & \multicolumn{3}{c|}{CelebAMask-HQ}  & \multicolumn{3}{c|}{CityScapes}  & \multicolumn{3}{c|}{ADE20K} & \Facades\\ \cmidrule{2-11}
{}     & mIoU  & accu  & FID           & mIoU  & accu  & FID      & mIoU  & accu  & FID      & FID    \\ \midrule
                       
Ground Truth  &73.14    &94.38   &9.41         &66.21   &93.69   &32.34       &39.38   &78.76  &14.51         &14.40 \\  \midrule    
Pix2PixHD~\cite{wang2018pix2pixHD}  &76.12    &95.76   &23.69         &50.35   & 92.09   &83.24       &22.78   &73.32  &43.0          &22.34 \\
SPADE~\cite{park2019SPADE}  &\bf{77.01}   &\bf{95.93}   &22.43         &56.01    & 93.13   &60.51       &\bf{35.37}   &\bf 79.37  &34.65       &24.04 \\
\bf{Ours} &75.69 &95.69 &\bf{17.66}   &\bf{57.88} &\bf{93.59} &\bf{50.38}     &34.59 &77.16 &\bf{24.84}  &\bf{19.82} \\
\bottomrule

\end{tabular}

\caption{Quantitative comparison using semantic segmentation performance measured by mIoU and accu and generation performance measured by FID. Our method outperforms current leading methods in FID on all the datasets.} 
\label{tab:SSM2Image quantitative table}
\end{table*}

\addtolength{\tabcolsep}{-4.5pt}    
\bgroup
\def\arraystretch{0.5}
\begin{figure*}[]
\begin{tabular} {cc|cc|c}
 Label & Ground Truth & Pix2PixHD~\cite{wang2018pix2pixHD} &  SPADE~\cite{park2019SPADE} & Ours\\

\includegraphics[width=0.1932\textwidth]{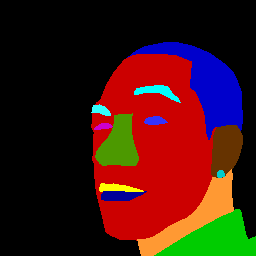} & \includegraphics[width=0.1932\textwidth]{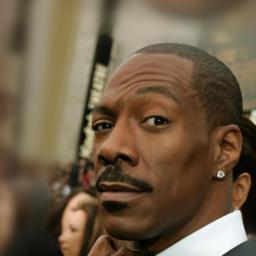} &
\includegraphics[width=0.1932\textwidth]{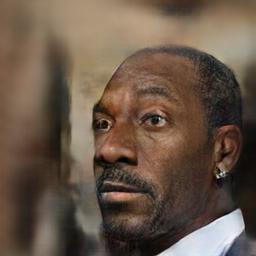}&
\includegraphics[width=0.1932\textwidth]{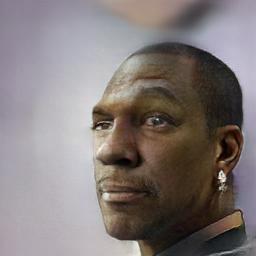} &  \includegraphics[width=0.1932\textwidth]{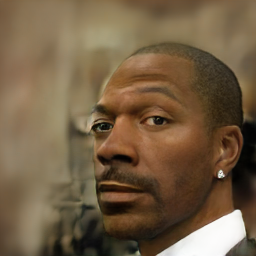} \\

\includegraphics[width=0.1932\textwidth]{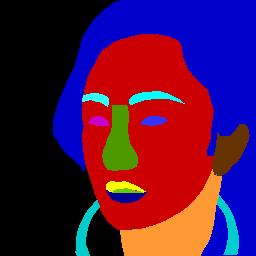} & \includegraphics[width=0.1932\textwidth]{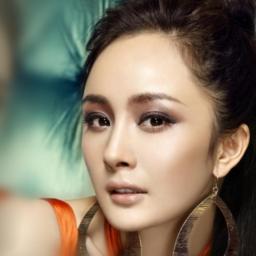} &
\includegraphics[width=0.1932\textwidth]{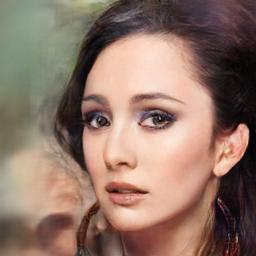}&
\includegraphics[width=0.1932\textwidth]{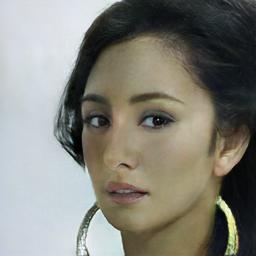} &  \includegraphics[width=0.1932\textwidth]{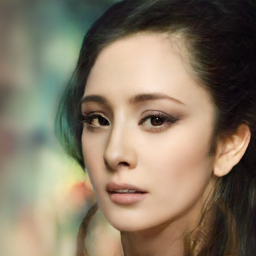} \\

\includegraphics[width=0.1932\textwidth,height=0.96in]{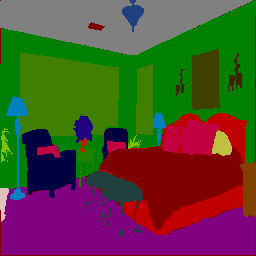} & \includegraphics[width=0.1932\textwidth,height=0.96in]{Images/Rec/ADE/gt/ADE_val_00000146.jpg} &
\includegraphics[width=0.1932\textwidth,height=0.96in]{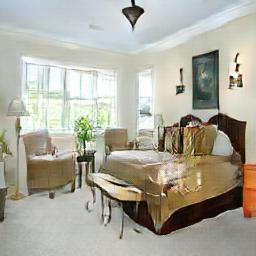} &   \includegraphics[width=0.1932\textwidth,height=0.96in]{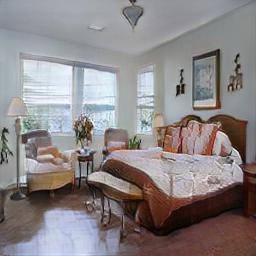} &  \includegraphics[width=0.1932\textwidth,height=0.96in]{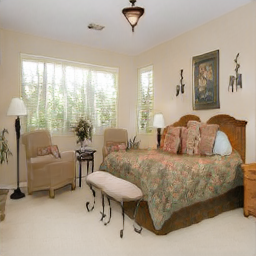} \\

\includegraphics[width=0.1932\textwidth,height=0.96in]{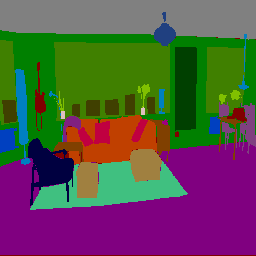} & \includegraphics[width=0.1932\textwidth,height=0.96in]{Images/Rec/ADE/gt/ADE_val_00000532.jpg} &
\includegraphics[width=0.1932\textwidth,height=0.96in]{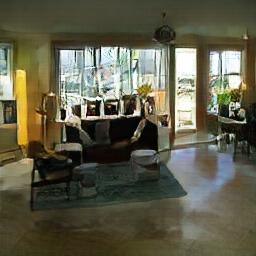} &   \includegraphics[width=0.1932\textwidth,height=0.80in]{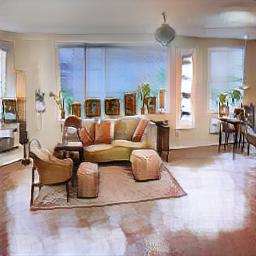} &  \includegraphics[width=0.1932\textwidth,height=0.96in]{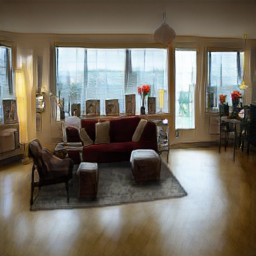} \\

\includegraphics[width=0.1932\textwidth,height=0.96in]{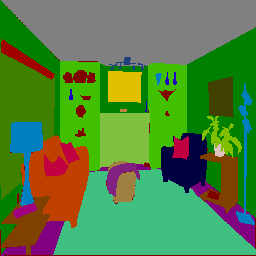} & \includegraphics[width=0.1932\textwidth,height=0.96in]{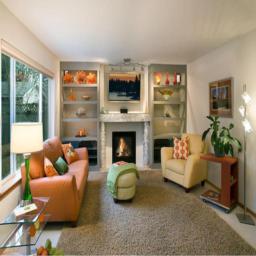} &
\includegraphics[width=0.1932\textwidth,height=0.96in]{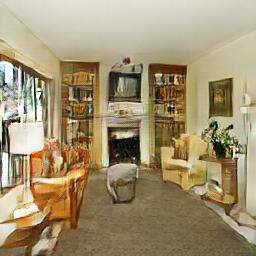} &   \includegraphics[width=0.1932\textwidth,height=0.96in]{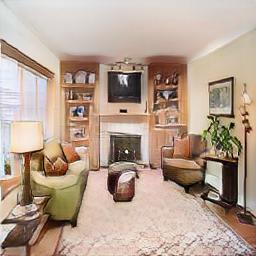} &  \includegraphics[width=0.1932\textwidth,height=0.96in]{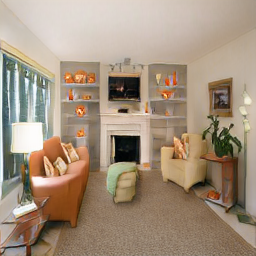} \\

\includegraphics[width=0.1932\textwidth]{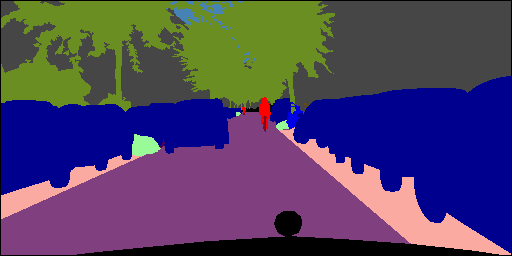} & \includegraphics[width=0.1932\textwidth]{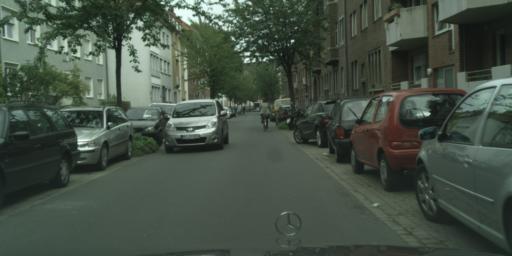} &
\includegraphics[width=0.1932\textwidth]{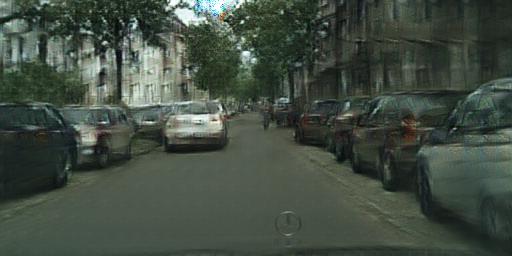} &   \includegraphics[width=0.1932\textwidth]{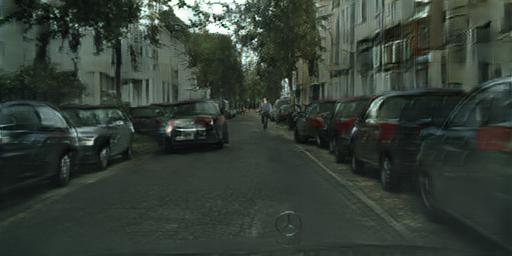} &  \includegraphics[width=0.1932\textwidth]{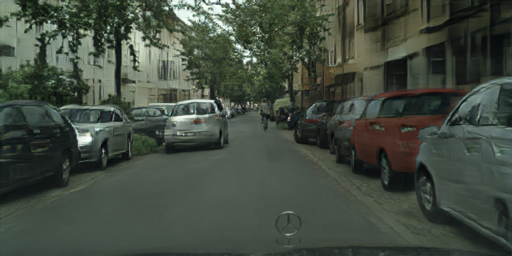} \\

\includegraphics[width=0.1932\textwidth]{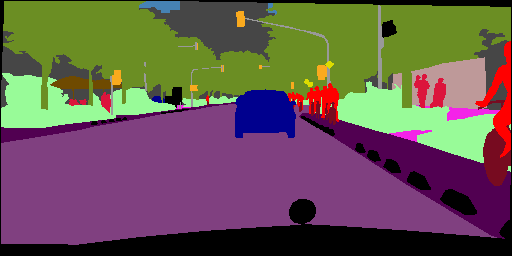} & \includegraphics[width=0.1932\textwidth]{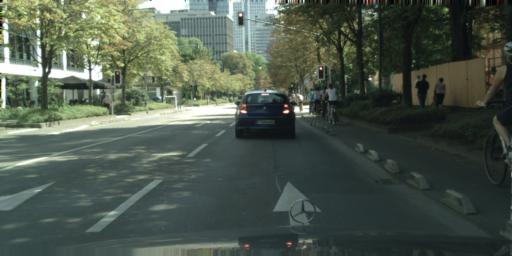} &
\includegraphics[width=0.1932\textwidth]{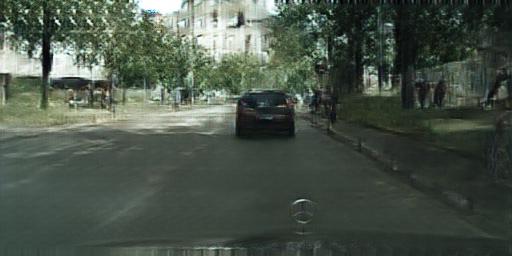} &   \includegraphics[width=0.1932\textwidth]{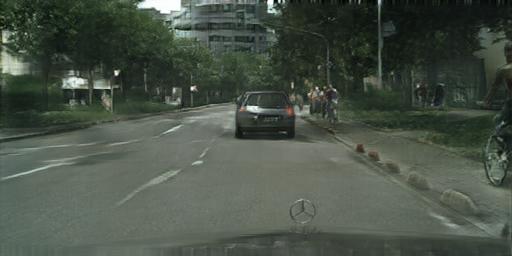} &  \includegraphics[width=0.1932\textwidth]{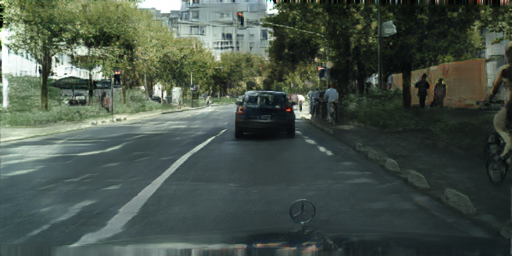} \\

\includegraphics[width=0.1932\textwidth]{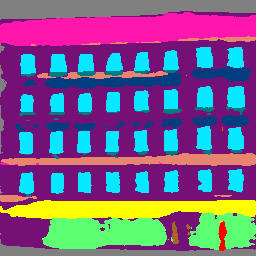} & \includegraphics[width=0.1932\textwidth]{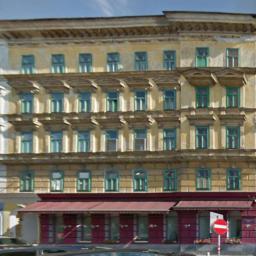} &
\includegraphics[width=0.1932\textwidth]{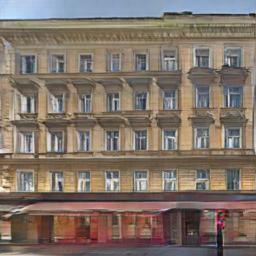}&
\includegraphics[width=0.1932\textwidth]{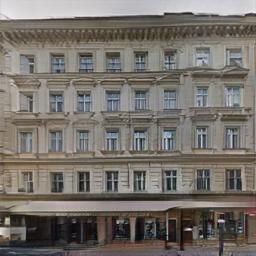} &  \includegraphics[width=0.1932\textwidth]{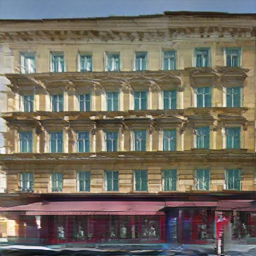} \\

\end{tabular}
\vspace{-2mm}
	\caption{Visual  comparison  of  semantic  image  synthesis  results  on  the  CelebAMask-HQ, ADE20K, CityScapes and \Facades dataset. We compare Pix2PixHD, SPADE, and our method. See supplementary materials for further details.}
	\label{fig:Reconstruction results}	
\vspace{-3mm}	
 \end{figure*}
 \egroup
 \addtolength{\tabcolsep}{4.5pt}

\begin{figure*}[th]
\centering
\includegraphics[width=\linewidth]{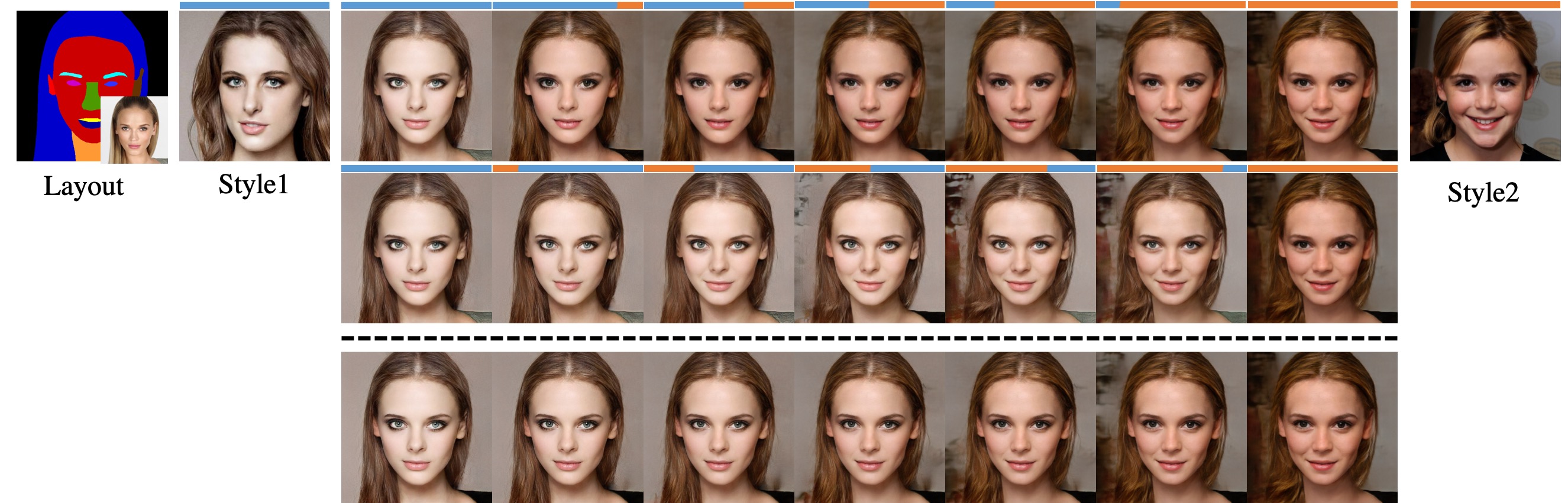}
\caption{Style crossover. In addition to style interpolation (bottom row), we can perform crossover by selecting different styles per ResBlk. We show two transitions in the top two rows. The blue / orange bars on top of the images indicate which styles are used by the six ResBlks. We can observe that earlier layers are responsible for larger features and later layers mainly determine the color scheme. 
}
\label{fig:style crossover}
\end{figure*}

\section{Experimental Setup}

\subsection{Network Architecture}

Figure \ref{fig:generator} (A) shows an overview of our generator network, which is building on that of SPADE~\cite{park2019SPADE}.
Similar to~\cite{park2019SPADE}, we employ a generator consisting of several SEAN ResNet blocks (SEAN ResBlk) with upsampling layers.

\vspace*{2mm}\noindent {\bf SEAN ResBlk.}
Figure~\ref{fig:generator} (B) shows the structure of our SEAN ResBlk, which consists of three convolutional layers whose scales and biases are modulated by three SEAN blocks respectively. 
Each SEAN block takes two inputs: a set of per-region style codes $\mathbf{ST}$ and a semantic mask $\mathbf{M}$.
Note that both inputs are adjusted in the beginning: the input segmentation mask is downsampled to the same height and width of the feature maps in a layer; the input style codes from $\mathbf{ST}$ are transformed per region using a $1 \times 1$ conv layer $A_{ij}$.
We observed that the initial transformations are indispensable components of the architecture because they transform the style codes according to the different roles of each neural network layer. 
For example, early layers might control the hair styles (\eg wavy, straight) of human face images while later layers might refer to the lighting and color.
In addition, we observed that adding noise to the input of SEAN can improve the quality of synthesized images.
The scale of such noise is adjusted by a per-channel scaling factor $B$ that is learnt during training, similar to StyleGAN~\cite{karras2018stylebased}.

\subsection{Training and Inference}

\vspace*{2mm}\noindent {\bf Training} We formulate the training as an image reconstruction problem. That is, the style encoder is trained to distill per-region style codes from the input images according to their corresponding segmentation masks.
The generator network is trained to reconstruct the input images with the extracted per-region style codes and the corresponding segmentation masks as inputs. 
Following SPADE~\cite{park2019SPADE} and Pix2PixHD~\cite{wang2018pix2pixHD}, the difference between input images and reconstructed images are measured by an overall loss function consisting of three loss terms: conditional adversarial loss, feature matching loss~\cite{wang2018pix2pixHD} and perceptual loss~\cite{johnson2016perceptual}. 
Details of the loss function are included in the supplementary material.

\vspace*{2mm}\noindent {\bf Inference}
During inference, we take an arbitrary segmentation mask as the mask input and implement per-region style control by selecting a separate $512$ dimensional style code for each semantic region as the style input.
This enables a variety of high quality image synthesis applications, which will be introduced in the following section.

\section{Results}

In the following we discuss quantitative and qualitative results of our framework.

\vspace*{2mm}\noindent {\bf Implementation details.} Following SPADE~\cite{park2019SPADE}, we apply Spectral Norm~\cite{miyato2018spectral} in both the generator and discriminator. Additional normalization is performed by SEAN in the generator. We set the learning rates to $0.0001$ and $0.0004$ for the generator and discriminator, respectively~\cite{heusel2017gans}. For the optimizer, we choose ADAM~\cite{kingma2014adam} with $\beta_{1}=0, \beta_{2}=0.999$. All the experiments are trained on $4$ Tesla v100 GPUs. To get better performance, we use a synchronized version of batch normalization~\cite{Zhang_2018_CVPR} in the SEAN normalization blocks.

\vspace*{2mm}\noindent {\bf Datasets.} We use the following datasets in our experiments: 1) CelebAMask-HQ~\cite{CelebAMask-HQ,karras2017progressive,liu2015faceattributes} containing $30,000$ segmentation masks for the CelebAHQ face image dataset. There are 19 different region categories.
2) ADE20K~\cite{zhou2017scene} contains $22,210$ images annotated with 150 different region labels.
3) Cityscapes~\cite{Cordts2016Cityscapes} contains 3500 images annotated with 35 different region labels. 
4) \Facades dataset. we use a dataset of $30,000$ \facade images collected from Google Street View~\cite{anguelov2010google}.  Instead of manual annotation, the segmentation masks are automatically calculated by a pre-trained DeepLabV3+ Network~\cite{chen2018encoderdecoder}.

We follow the recommended splits for each dataset and note that all images in the paper are generated from the test set only. Neither the segmentation masks nor the style images have been seen during training.

\begin{table}[hpb]
\centering
\small
\begin{tabular}{lcccc} \toprule

Method                          &SSIM    &RMSE  &PSNR    &FID \\ \midrule
                       
Pix2PixHD~\cite{wang2018pix2pixHD}                       &0.68   &0.15   &17.14   &23.69\\
SPADE~\cite{park2019SPADE}                           &0.63   &0.21   &14.30   &22.43\\
SPADE++                         &0.67   &0.16   &16.71   &20.80\\ \midrule
SEAN-level encoders    &0.74   &0.11   &19.70   &18.17\\
ResBlk-level encoders           &0.72   &0.12   &18.86   &17.98\\
\bf{unified encoder}        &0.73   &0.12   &18.74   &17.66\\ \midrule
\bf{w/ downsampling}&0.73   &0.12   &18.74   &17.66\\ 
w/o downsampling        &0.72   &0.13   &18.32   &18.67\\ \midrule
\bf{w/ noises}              &0.73   &0.12   &18.74   &17.66\\
w/o noises                      &0.72   &0.13   &18.49   &18.05\\
\bottomrule

\end{tabular}
\caption{Ablation study on CelebAMask-HQ dataset. See supplementary materials for further details.} 
\label{tab:ablation study table}
\end{table}

\vspace*{2mm}\noindent {\bf Metrics.} We employ the following established metrics to compare our results to state of the art: 1) segmentation accuracy measured by \ac{mIoU} and \ac{accu}, 2) FID~\cite{heusel2017gans}, 3) peak signal-to-noise ratio (PSNR), 4) structural similarity (SSIM)~\cite{Wang04imagequality}, 5) root mean square error (RMSE).

\vspace*{2mm}\noindent {\bf Quantitative comparisons.} 
In order to facilitate a fair comparison to SPADE, we report reconstruction performance where only a single style image is employed.
We train one network for each of the four datasets and report the previously described metrics in Table~\ref{tab:similarity quantitative table} and~\ref{tab:SSM2Image quantitative table}. We choose SPADE~\cite{park2019SPADE} as the currently best state-of-the-art method and Pix2PixHD~\cite{wang2018pix2pixHD} as the second best method in the comparisons. Based on visual inspection of the results, we found that the FID score is most indicative of visual quality of the results and that a lower FID score often (but not always) corresponds to better visual results. Generally, we can observe that our SEAN framework clearly beats the state of the art on all datasets.

\vspace*{2mm}\noindent {\bf Qualitative results.}
We show visual comparisons for the four datasets in Fig.~\ref{fig:Reconstruction results}. We can observe that the quality difference between our work and the state of the art can be significant. For example, we notice that SPADE and Pix2Pix cannot handle more extreme poses and older people. We conjecture that our network is more suitable to learn more of the variability in the data because of our improved style encoding.
Since a visual inspection of the results is very important for generative modeling, we would like to refer the reader to the supplementary materials and the accompanying video for more results. Our per-region style encoding also enables new image editing operations: iterative image editing with per-region style control (See Figs.~\ref{fig:teaser} and \ref{fig:ADE editing}), style interpolation and style crossover (See Fig.~\ref{fig:style interpolation} and \ref{fig:style crossover}).

\vspace*{2mm}\noindent {\bf Variations of SEAN generator (Ablation studies).}
In Table~\ref{tab:ablation study table}, we compare many variations of our architecture with previous work on the CelebAMask-HQ dataset.

According to our analysis, Pix2PixHD has a better style extraction subnetwork than SPADE (because the PSNR reconstruction values are better), but SPADE has a better generator subnetwork (because the FID scores are better). We therefore build another network that combines the style encoder from Pix2PixHD with the generator from SPADE and call this architecture SPADE++. We can observe that SPADE++ improves SPADE by a bit, but all our architecture variations still have a better FID and PSNR score.

To evaluate our design choices, we report the performance of variations of our encoder. First we compare three variants to encode style: SEAN-level encoder, ResBlk-level encoder, and unified encoder.
A fundamental design choice is how to split the style computation into two parts: a shared computation in the style encoder and a per layer computation in each SEAN block.
The first two variants perform all style extraction per-layer, while the unified encoder is the architecture described in the main part of the paper.
While the other two encoders have better reconstruction performance, they lead to lower FID scores. 
Based on visual inspection, we also confirmed that the unified encoder leads to better visual quality. This is especially noticeable for difficult inputs (see supplementary materials).
Second, we evaluate if downsampling in the style encoder is a good idea and evaluate a style encoder with bottleneck (with downsampling) with another encoder that uses less downsampling. This test confirms that introducing a bottleneck in the style encoder is a good idea. Lastly, we test the effect of the learnable noise. We found the learnable noise can help both similarity (PSNR) and FID performance. More details of the alternative architectures are provided in supplementary materials. 


\section{Conclusion}

We propose semantic region-adaptive normalization (SEAN), a simple but effective building block for Generative Adversarial Networks (GANs) conditioned on segmentation masks that describe the semantic regions in the desired output image.
Our main idea is to extend SPADE, the currently best network, to control the style of each semantic region individually, e.g. we can specify one style reference image per semantic region. We introduce a new building block, SEAN normalization, that can extract style from a given style image for a semantic region and processes the style information to bring it in the form of spatially-varying normalization parameters.
We evaluate SEAN on multiple datasets and report better quantitative metrics (e.g. FID, PSNR) than the current state of the art.
While SEAN is designed to specify style for each region independently, we also achieve big improvements in the basic setting where only one style image is provided for all regions.
In summary, SEAN is better suited to encode, transfer, and synthesize style than the best previous method in terms of reconstruction quality, variability, and visual quality. SEAN also pushes the frontier of interactive image editing. In future work, we plan to extend SEAN to processing meshes, point clouds, and textures on surfaces.

\textbf{Acknowledgement} This work was supported by the KAUST Office of Sponsored Research (OSR) under Award No. OSR-CRG2018-3730.


{\small
\bibliographystyle{ieee_fullname}
\bibliography{egbib}
}
\clearpage
\clearpage
\newpage
\appendix

\section{Additional Implementation Details}

\begin{figure}[ht]
\centering
\includegraphics[width=0.9\linewidth]{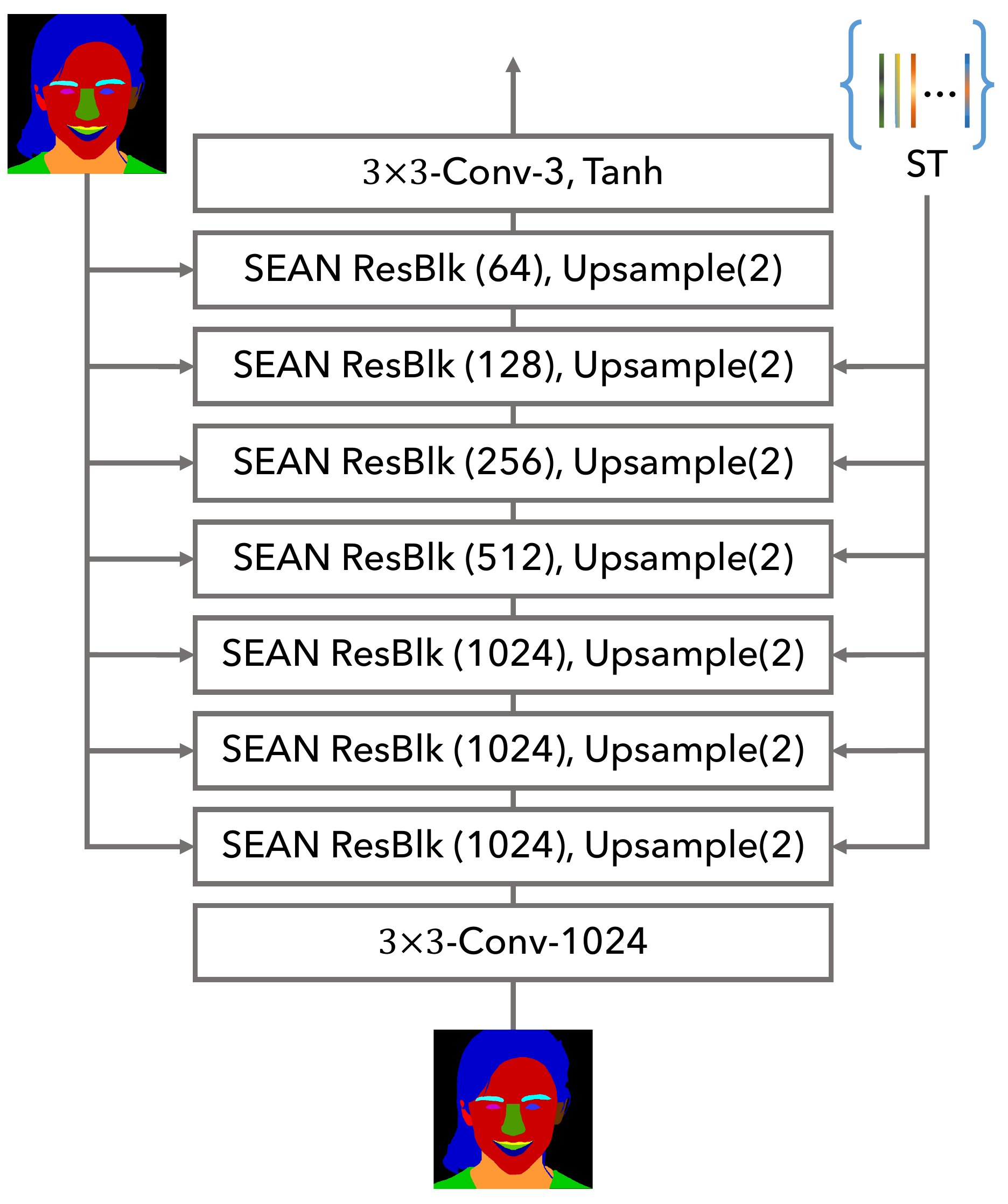}
\caption{SEAN Generator. The style codes $\mathbf{ST}$ and segmentation mask are passed to the generator through the proposed SEAN ResBlks. The number of feature map channels is shown in the parenthesis after each SEAN ResBlk. To better illustrate the architecture, we omit the learnable noise inputs and per-style Conv layers in this figure. These details are shown in Fig.3 of the main paper (see $A_{ij}$ and $B_{ij}$).}
\label{fig:SEAN generator architecture}
\end{figure}

\noindent {\bf Generator.} Our generator consists of several SEAN ResBlks. Each of them is followed by a nearest neighbor upsampling layer. Note that we only inject the style codes $\mathbf{ST}$ into the first 6 SEAN ResBlks. The other inputs are injected to all SEAN ResBlks. The architecture of our generator is shown in Figure~\ref{fig:SEAN generator architecture}.

\begin{figure}[ht]
\centering
\includegraphics[width=0.9\linewidth]{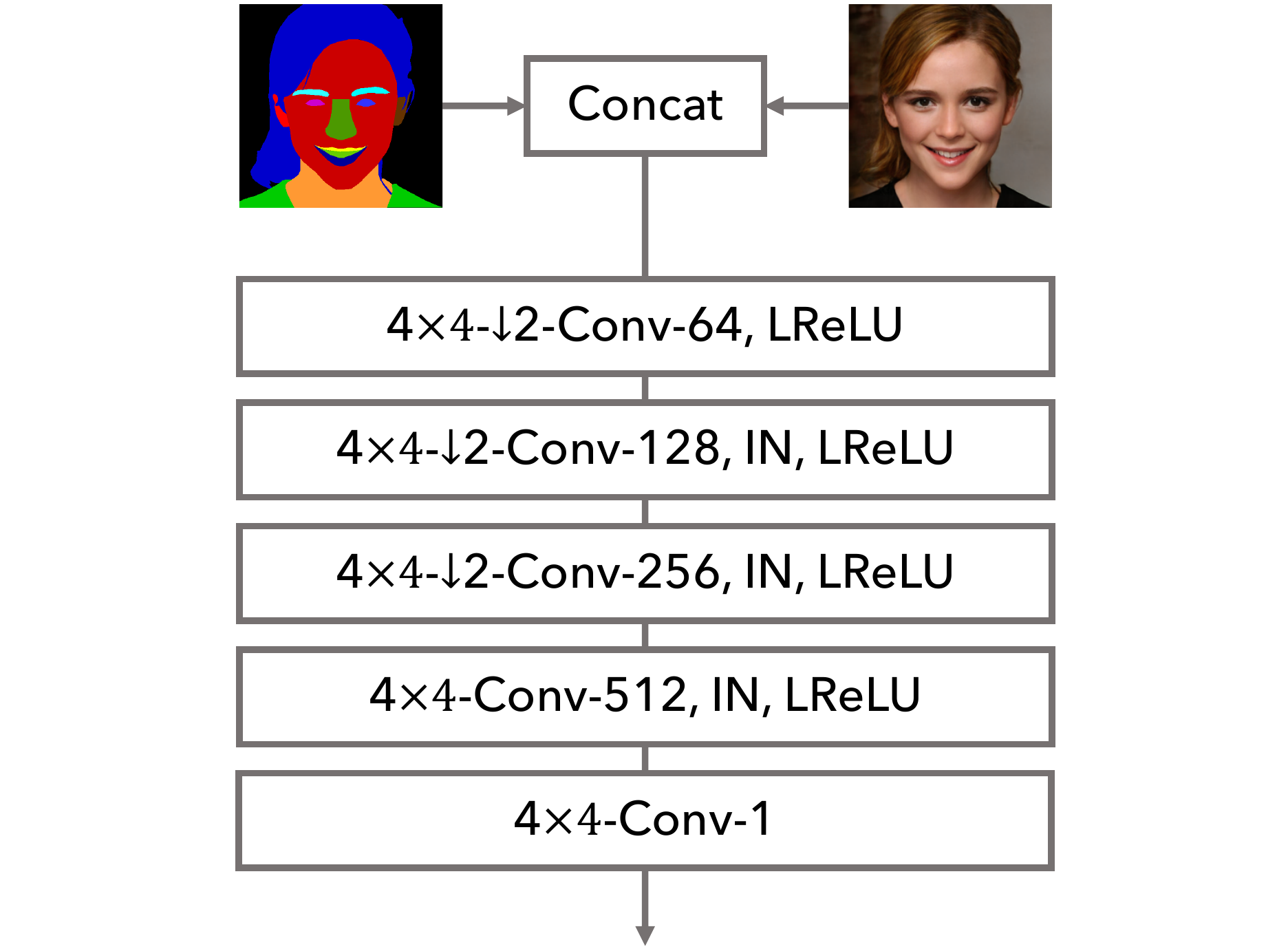}
\caption{Following SPADE and Pix2PixHD, our discriminator takes the concatenation of a segmentation mask and a style image as inputs. The loss is calculated in the same way  as PatchGAN~\cite{isola2016imagetoimage}.}
\label{fig:SEAN discriminator architecture}
\end{figure}

\noindent {\bf Discriminator.} Following SPADE~\cite{park2019SPADE} and Pix2PixHD~\cite{wang2018pix2pixHD}, we employed two multi-scale discriminators with instance normalization (IN)~\cite{ulyanov2016instance} and Leaky ReLU (LReLU). Similar to SPADE, we apply spectral normalization~\cite{miyato2018spectral} to all the convolutional layers of the discriminator. The architecture of our discriminator is shown in Figure~\ref{fig:SEAN discriminator architecture}.

\noindent {\bf Style Encoder.} Our style encoder consists of a ``bottleneck'' convolutional neural network and a region-wise average pooling layer (Figure~\ref{fig:SEAN style encoder architecture}). The inputs are the style image and the corresponding segmentation mask, while the outputs are the style codes $\mathbf{ST}$.

\begin{figure*}[th]
\centering
\includegraphics[width=\linewidth]{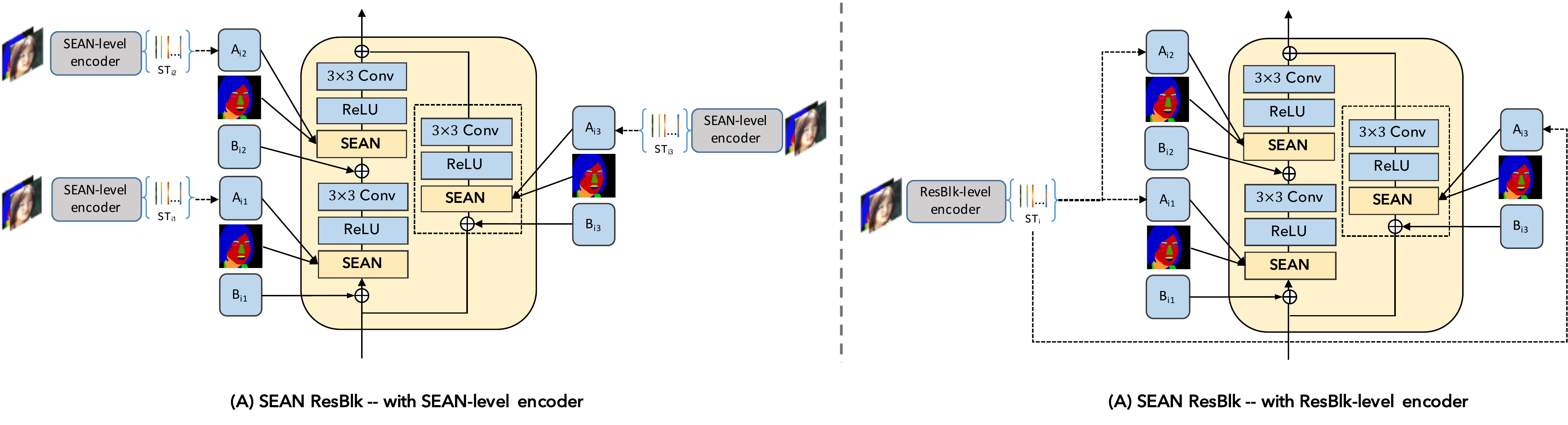}
\caption{Detailed usage of the SEAN-level encoder and the ResBlk-level encoder within a SEAN ResBlk.}
\label{fig:different encoders}
\end{figure*}

\begin{figure}[ht]
\centering
\includegraphics[width=0.9\linewidth]{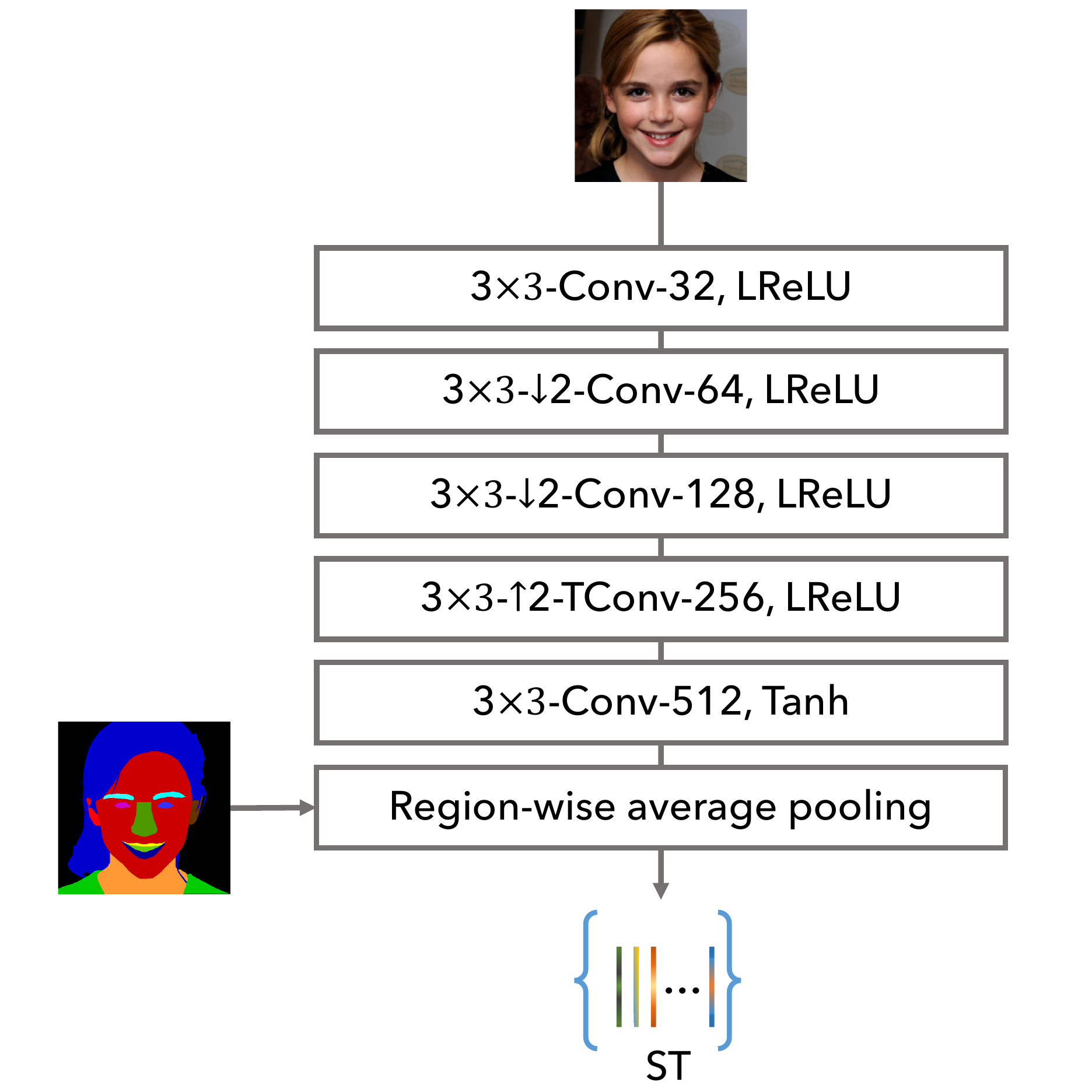}
\caption{Our style encoder takes the style image and the segmentation mask as inputs to generate the style codes $\mathbf{ST}$.}
\label{fig:SEAN style encoder architecture}
\end{figure}

\vspace*{3mm}\noindent {\bf Loss function.} The design of our loss function is inspired by those of SPADE and Pix2PixHD which contains three components:\\
 
\noindent (1) \textit{Adversarial loss.}
Let $E$ be the style encoder, $G$ be the SEAN generator, $D_1$ and $D_2$ be two discriminators at different scales~\cite{wang2018pix2pixHD}, $\mathbf{R}$ be a given style image, $\mathbf{M}$ be the corresponding segmentation mask of $\mathbf{R}$, we formulate the conditional adversarial learning part of our loss function as:
\small
\begin{equation}
\min _{E, G} \max _{D_{1}, D_{2}} \sum_{k=1,2} \mathcal{L}_{\mathrm{GAN}}\left(E, G, D_{k}\right)
\end{equation}
\normalsize
Specifically, $\mathcal{L}_{\mathrm{GAN}}$ is built with the Hinge loss that:
\small
\begin{equation}
\begin{split}
\mathcal{L}_{\mathrm{GAN}} &= \mathbb{E}\left[\max(0, 1-D_{k}(\mathbf{R, M}))\right] \\
&+  \mathbb{E}\left[\max(0, 1+D_{k}(G\mathbf{\left(\mathbf{ST,M}\right),M}))\right]
\label{eq:adversarial loss function}
\end{split}
\end{equation}
\normalsize
where $\mathbf{ST}$ is the style codes of $\mathbf{R}$ extracted by $E$ under the guidance of $\mathbf{M}$:
\small
\begin{equation}
\mathbf{ST} = E\left(\mathbf{R, M}\right)
\label{eq:ST function}
\end{equation}
\normalsize

\noindent (2) \textit{Feature matching loss}~\cite{wang2018pix2pixHD}. Let $T$ be the total number of layers in discriminator $D_{k}$, $D_{k}^{(i)}$ and $N_{i}$ be the output feature maps and the number of elements of the $i$-th layer of $D_{k}$ respectively, we denote the feature matching loss term $\mathcal{L}_{FM}$ as:

\small
\begin{equation}
\mathcal{L}_{FM}=\mathbb{E} \sum_{i=1}^{T}\frac{1}{N_{i}}\left[\left\|D_{k}^{(i)}(\mathbf{R} ,\mathbf{M})-D_{k}^{(i)}\left(G\mathbf{(ST,M)},\mathbf{M}\right)\right\|_{1}\right]
\label{eq:feature matching loss function}
\end{equation}
\normalsize

\noindent (3) \textit{Perceptual loss}~\cite{johnson2016perceptual}. Let $N$ be the total number of layers used to calculate the perceptual loss, $F^{(i)}$ be the output feature maps of the $i$-th layer of the VGG network~\cite{simonyan2014deep}, $M_{i}$ be the number of elements of $F^{(i)}$, we denote the perceptual loss $\mathcal{L}_{p e r c e p t}$ as:
\small
\begin{equation}
\mathcal{L}_{p e r c e p t}=\mathbb{E}\sum_{i=1}^{N} \frac{1}{M_{i}}\left[\left\|F^{(i)}\left(\mathbf{R}\right)-F^{(i)}\left(G\mathbf{(ST,M)}\right)\right\|_{1}\right]
\label{eq:perceptual loss function}
\end{equation}
\normalsize

The final loss function used in our experiment is made up of the above-mentioned three loss terms as: 
\small
\begin{equation}
\min_{E, G}\bigg( \Big( \max _{D_{1}, D_{2}} \sum_{k=1,2} \mathcal{L}_{\mathrm{GAN}} \Big) + \lambda_{1} \sum_{k=1,2} \mathcal{L}_{\mathrm{FM}} + \lambda_{2} \mathcal{L}_{\mathrm{p e r c e p t}}\bigg) 
\label{eq:overall loss}
\end{equation}
\normalsize
Following SPADE and Pix2PixHD, we set $\lambda_{1} = \lambda_{2} = 10$ in our experiments.

\vspace*{3mm}\noindent {\bf Training details.} We perform $50$ epochs of training on all the datasets mentioned in the main paper. During training, all input images are resized to a resolution of $256\times256$, except for the CityScapes dataset~\cite{Cordts2016Cityscapes} whose images are resized to $512\times256$. We use Glorot initialization~\cite{pmlr-v9-glorot10a} to initialize our network weights.

\section{Additional Experimental Details}

\noindent \textbf{Table 3 (main paper).}
Supplementing row 5 and 6 in Table 3 of the main paper, Figure~\ref{fig:different encoders} shows how the two variants of style encoders (\ie the SEAN-level encoder and the ResBlk-level encoder) are used in a SEAN ResBlk.
Specifically, the SEAN-level encoders extract different style codes for each SEAN block while the same style codes extracted by the ResBlk-level encoder are used by all SEAN blocks within a SEAN ResBlk. 

\vspace*{3mm}\noindent \textbf{Figure 6 (main paper).}
We used the Ground Truth (second column in Figure 6 of the main paper) as the style input for all methods.
For Pix2pixHD, we generate the results by: (i) encoding the style image into a style vector; (ii) broadcasting the style vector and concatenating it to the mask input of the generator.

\section{Justification of Encoder Choice} Figure~\ref{fig:Encoder Comparison} shows that the images generated by the unified encoder are of better visual quality than those generated by the SEAN-level encoder, especially for challenging inputs (e.g. extreme poses, unlabeled regions), which justifies our choice of unified encoder.

\addtolength{\tabcolsep}{-4.5pt}    
\bgroup
\def\arraystretch{0.5}
\begin{figure}[th]
\small
\begin{tabular} {cc|cc}
Label & Style Image & Encoder1 & Encoder2
\\
\includegraphics[width=0.11\textwidth]{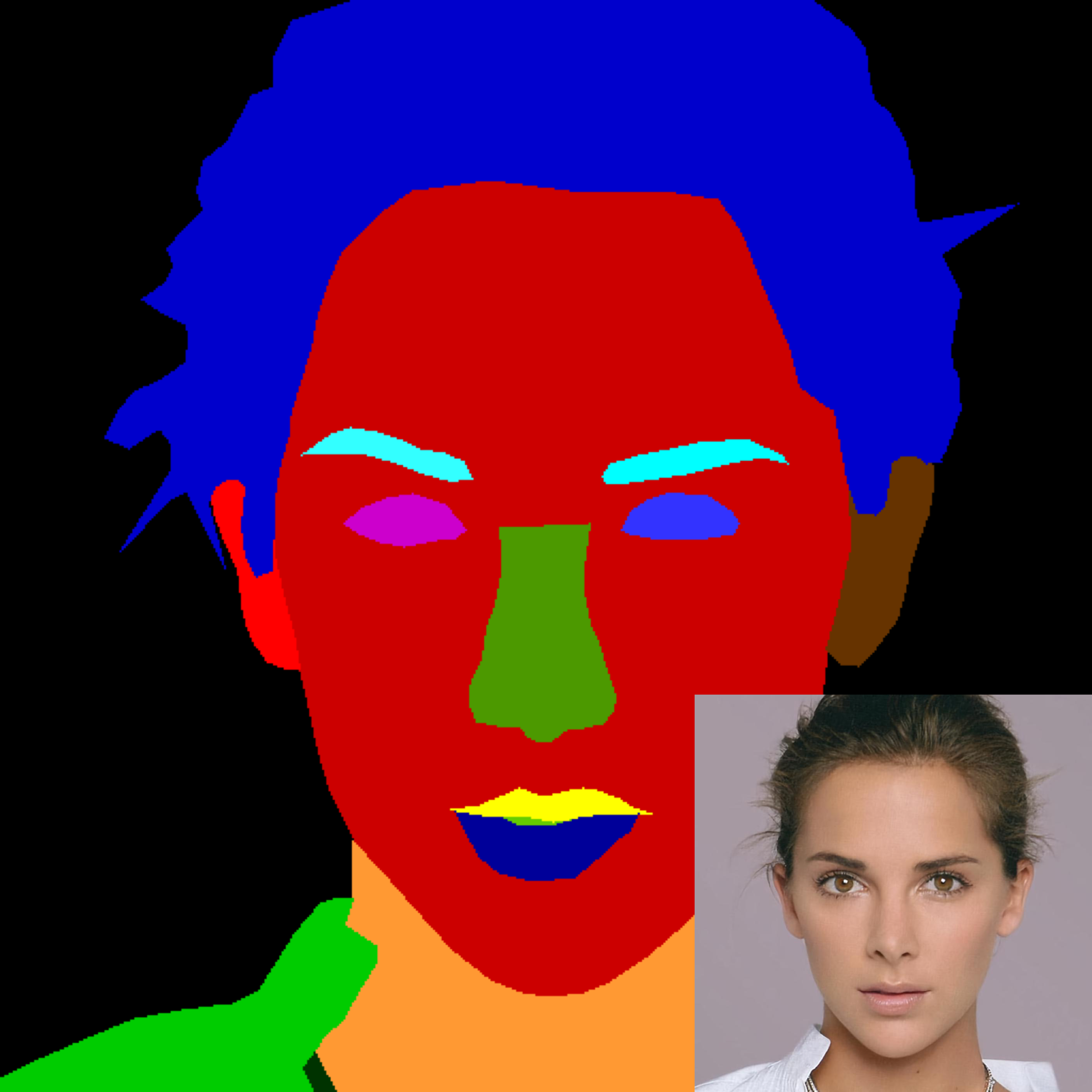} & \includegraphics[width=0.11\textwidth]{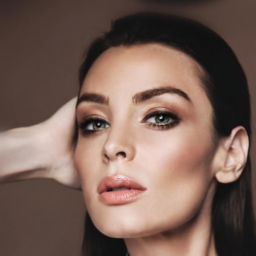} &
\includegraphics[width=0.11\textwidth]{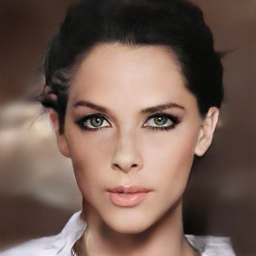} &   \includegraphics[width=0.11\textwidth]{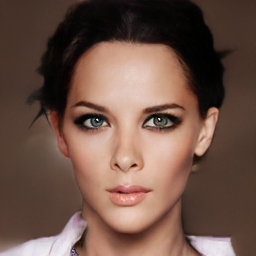}\\ 

\includegraphics[width=0.11\textwidth]{Images/Encoder_search/encoder_1.png} & \includegraphics[width=0.11\textwidth]{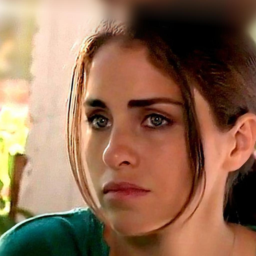} &
\includegraphics[width=0.11\textwidth]{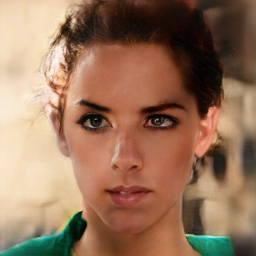} &   \includegraphics[width=0.11\textwidth]{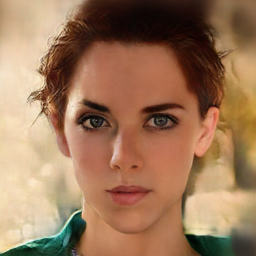}\\ 

\includegraphics[width=0.11\textwidth]{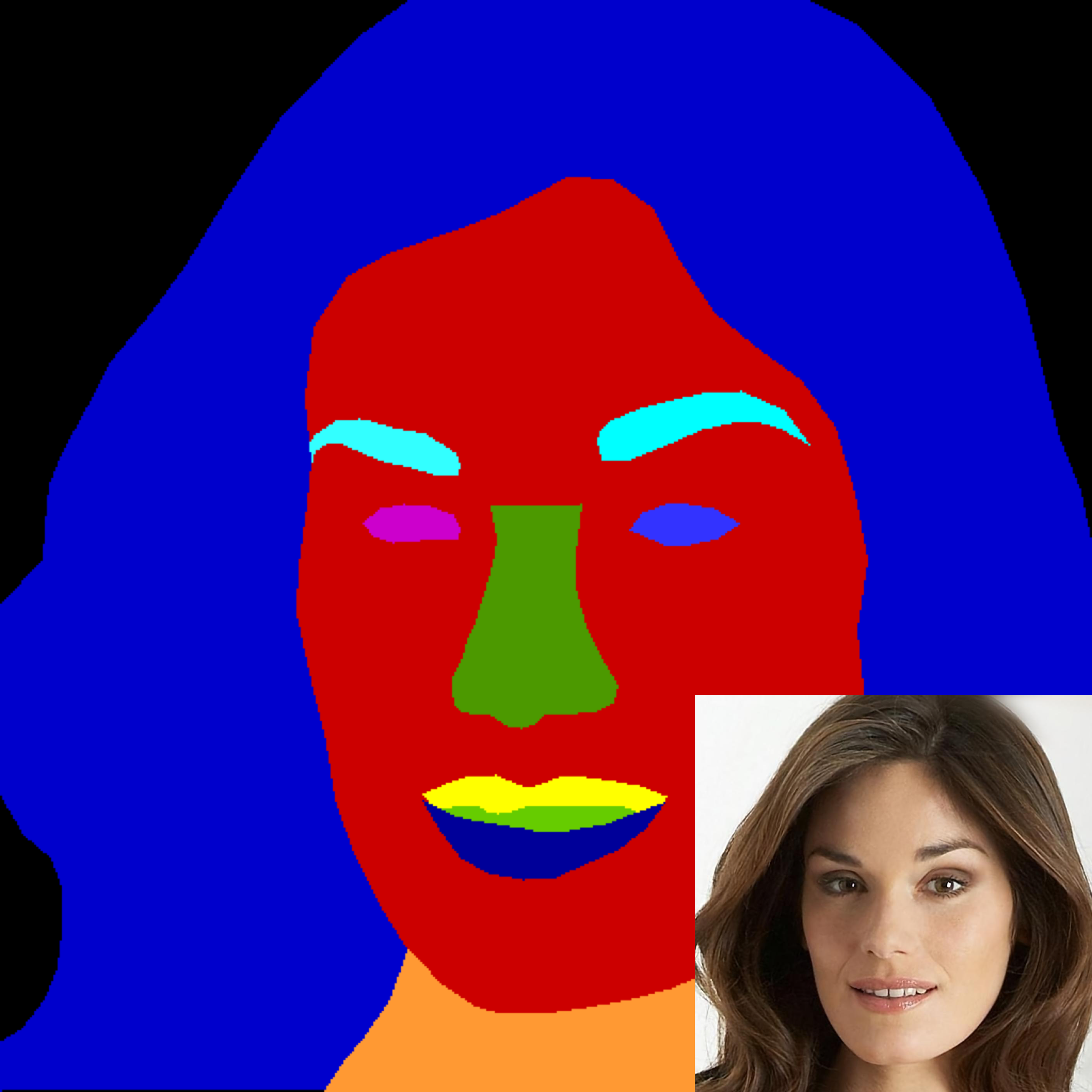} & \includegraphics[width=0.11\textwidth]{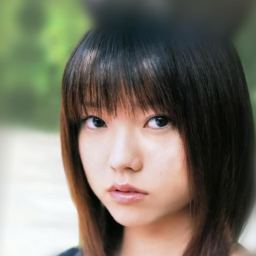} &
\includegraphics[width=0.11\textwidth]{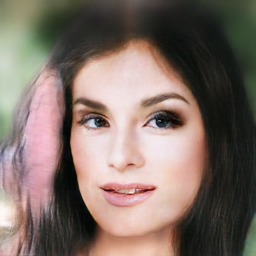} &   \includegraphics[width=0.11\textwidth]{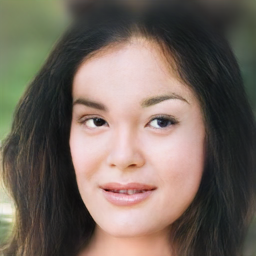}\\ 

\includegraphics[width=0.11\textwidth]{Images/Encoder_search/encoder_2.png} & \includegraphics[width=0.11\textwidth]{Images/Encoder_search/style/29659.png} &
\includegraphics[width=0.11\textwidth]{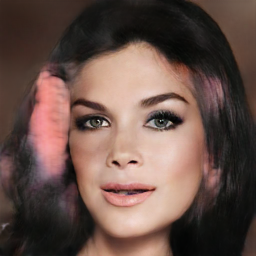} &   \includegraphics[width=0.11\textwidth]{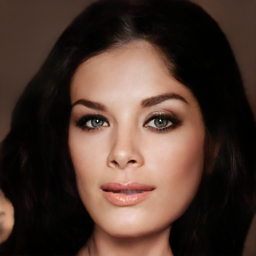}\\

\end{tabular}
\vspace{-2mm}
	\caption{Encoder choice justification. Encoder1 is the SEAN-level encoder and Encoder2 is the unified encoder. SEAN-level encoder is more sensitive to the poses and unlabeled parts of the style image due to the overfitting. Using unified encoder can get more robust style transfer results.}
	\label{fig:Encoder Comparison}	
\vspace{-3mm}	
 \end{figure}
 \egroup
 \addtolength{\tabcolsep}{4.5pt}

\section{Additional Analysis}
\begin{figure}[h]
\centering
\includegraphics[width=\linewidth]{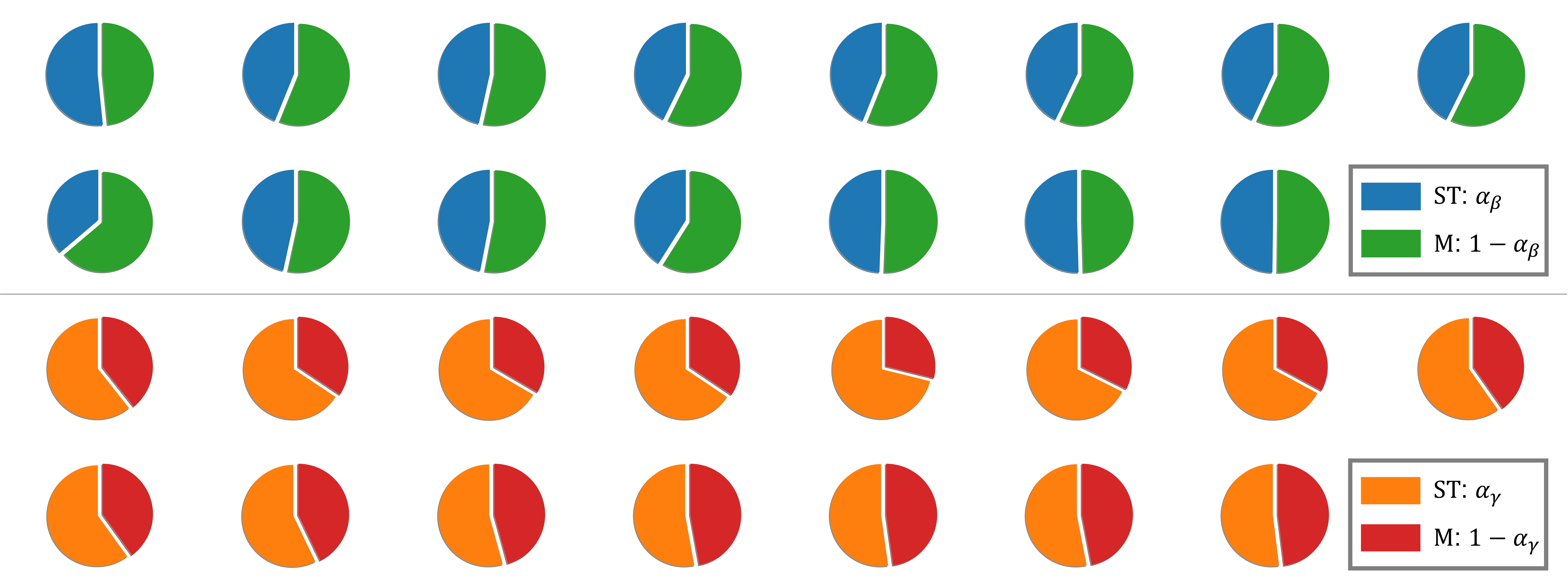}
\caption{Contributions of ST-branch and Mask-branch for each SEAN normalization block. The pie charts and SEAN normalization blocks are in one-to-one correspondence.}
\label{fig:weight histogram}
\end{figure}

\noindent {\bf ST-branch \vs Mask-branch.}
The contributions of ST-branch and mask-branch are determined by a linear combination (parameters $\alpha_\beta$ and $\alpha_\gamma$). The resulting parameters are typically in the range of $0.35 - 0.7$ meaning that both branches are actively contributing to the result. See Fig~\ref{fig:weight histogram} for one example.
It is possible to completely drop the mask-branch, but the results will get worse. 
It was our initial intuition that the mask branch provides the rough structure and the ST-branch additional details. However, in the end, the interaction is quite complicated and cannot be understood by just varying the mixing parameter.


\begin{figure}[h]
\centering
\includegraphics[width=\linewidth]{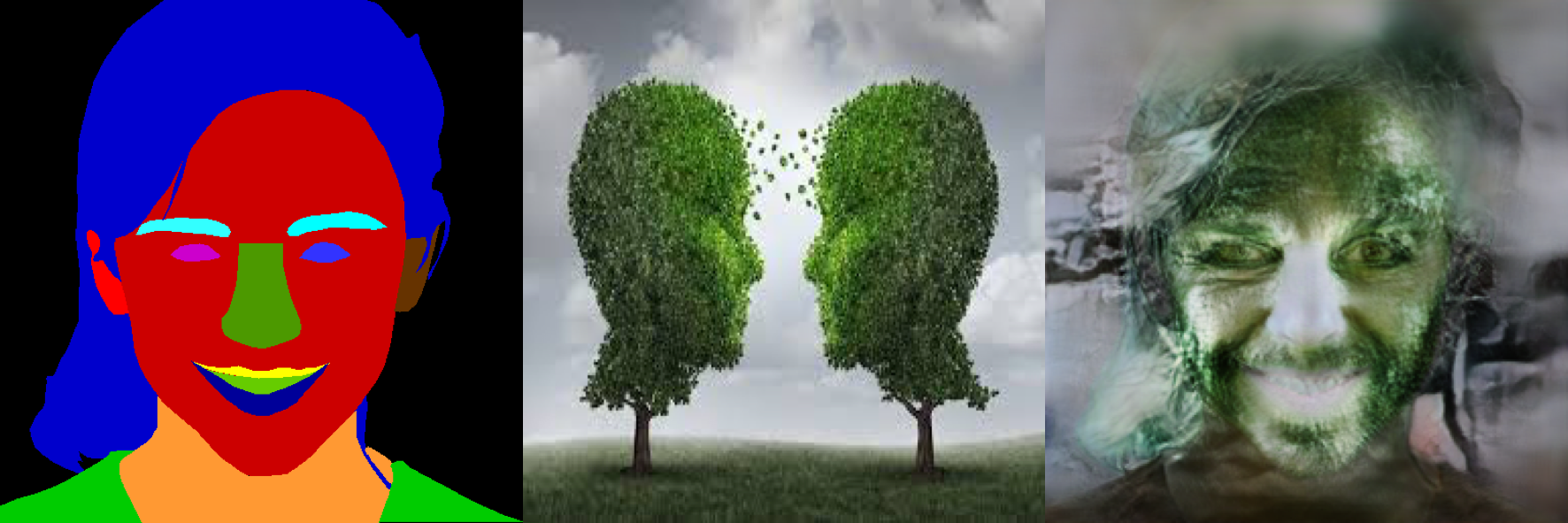}
\caption{Complex texture transfer.}
\label{fig:tree}
\end{figure}

\noindent \textbf{Extreme Cases.}
To further demonstrate SEAN's power in texture transfer, we show that highly complex textures from an artistic image can be transferred to a human face (Fig~\ref{fig:tree}).

\begin{figure}[h]
\centering
\includegraphics[width=1.0\linewidth]{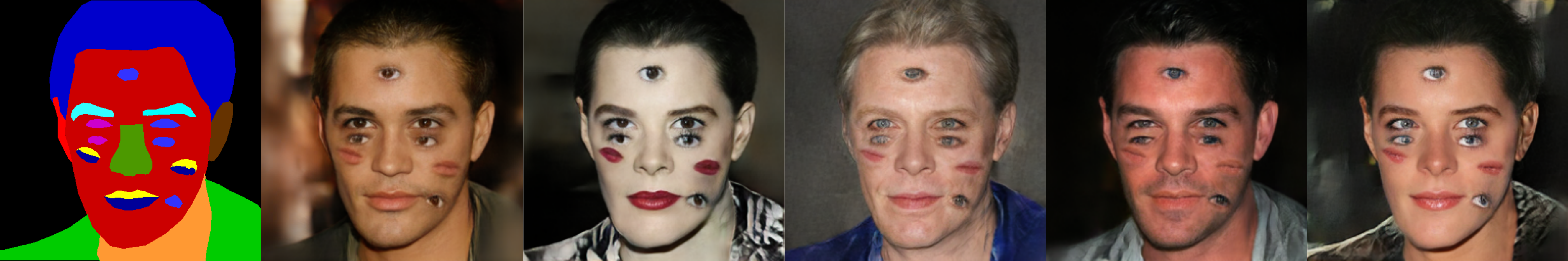}
\caption{Spatially-flexible painting. Our method allows users to put eyes anywhere on a face.}
\label{fig:monster}
\end{figure}

\vspace*{-5mm}
\noindent In addition, our method is highly flexible that enables users to paint a semantic region at a spatially unreasonable location arbitrarily (Fig~\ref{fig:monster}).

\section{User Study} 
We conducted a user preference study with Amazon Mechanical Turk (AMT) to illustrate our superior reconstruction results against existing methods (Table~\ref{tab:user study}).
Specifically, we created 600 questions for AMT workers to answer.
In the end, our questions are answered by 575 AMT workers.
For each question, we show the user a set of 5 images: a ground truth image, its corresponding segmentation mask, and 3 reconstruction images obtained by our method, Pix2PixHD~\cite{wang2018pix2pixHD} and SPADE~\cite{park2019SPADE}. The user is then asked to select the reconstructed image closest to the ground truth and with fewest artifacts.
To relieve the impact of image orders and make a fair comparison, we picked 100 image sets randomly and created the 600 questions by enumerating all the 6 possible orders of the 3 reconstructed images in each of them.

\begin{table}[thpb]
\centering
\small
\begin{tabular}{l|cccc} \toprule

                       

                          &Pix2PixHD~\cite{wang2018pix2pixHD}      &SPADE~\cite{park2019SPADE}  &Ours\\ \midrule
                       
Preference (\%)                    &23.17   &8.83  &68.00\\

\bottomrule

\end{tabular}
\caption{User preference study (CelebAMask-HQ dataset). Our method outperforms Pix2PixHD~\cite{wang2018pix2pixHD} and SPADE~\cite{park2019SPADE} significantly. 
} 
\label{tab:user study}
\end{table}

\section{Additional Results}

To demonstrate that the proposed per-region style control method builds the foundation of a highly flexible image-editing software, we designed an interactive UI for a demo.
Our UI enables high quality image synthesis by transferring the per-region styles from various images to an arbitrary segmentation mask.
New styles can be created by interpolating existing styles.
Please find the recorded videos of our demo in the supplementary material.

Figure~\ref{fig:additional style transfer} shows additional style transfer results on CelebAMask-HQ~\cite{CelebAMask-HQ,karras2017progressive,liu2015faceattributes} dataset. Figure~\ref{fig:Supp face interpolation} and Figure~\ref{fig:Supp ADE interpolation} show additional style interpolation results on CelebAMask-HQ and ADE20K datasets.

Figure~\ref{fig:CelebAMask-HQ results}, ~\ref{fig:ADE20K results}, ~\ref{fig:ADE20K outdoor results} and ~\ref{fig:City Facades results} show additional image reconstruction results of our method, Pix2PixHD and SPADE on the CelebAMask-HQ~\cite{CelebAMask-HQ,karras2017progressive,liu2015faceattributes}, ADE20K~\cite{zhou2017scene}, CityScapes~\cite{Cordts2016Cityscapes} and our \Facades datasets respectively.
It can be observed that our reconstructions are of much higher quality than those of Pix2PixHD and SPADE.

\begin{figure*}[th]
\centering
\includegraphics[width=\linewidth]{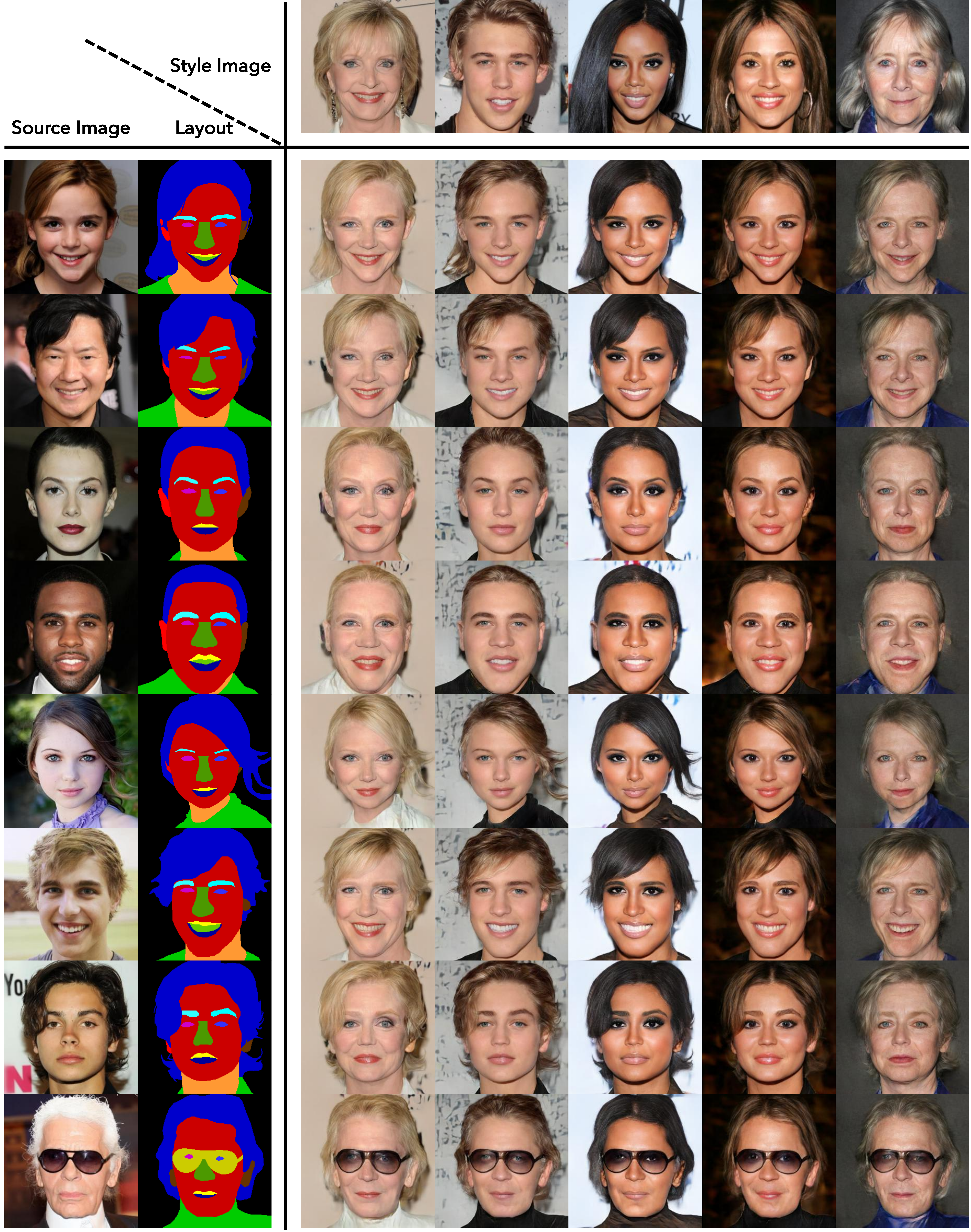}
\caption{Style transfer on CelebAMask-HQ dataset}
\label{fig:additional style transfer}
\end{figure*}

\begin{figure*}[th]
\centering
\includegraphics[width=\linewidth]{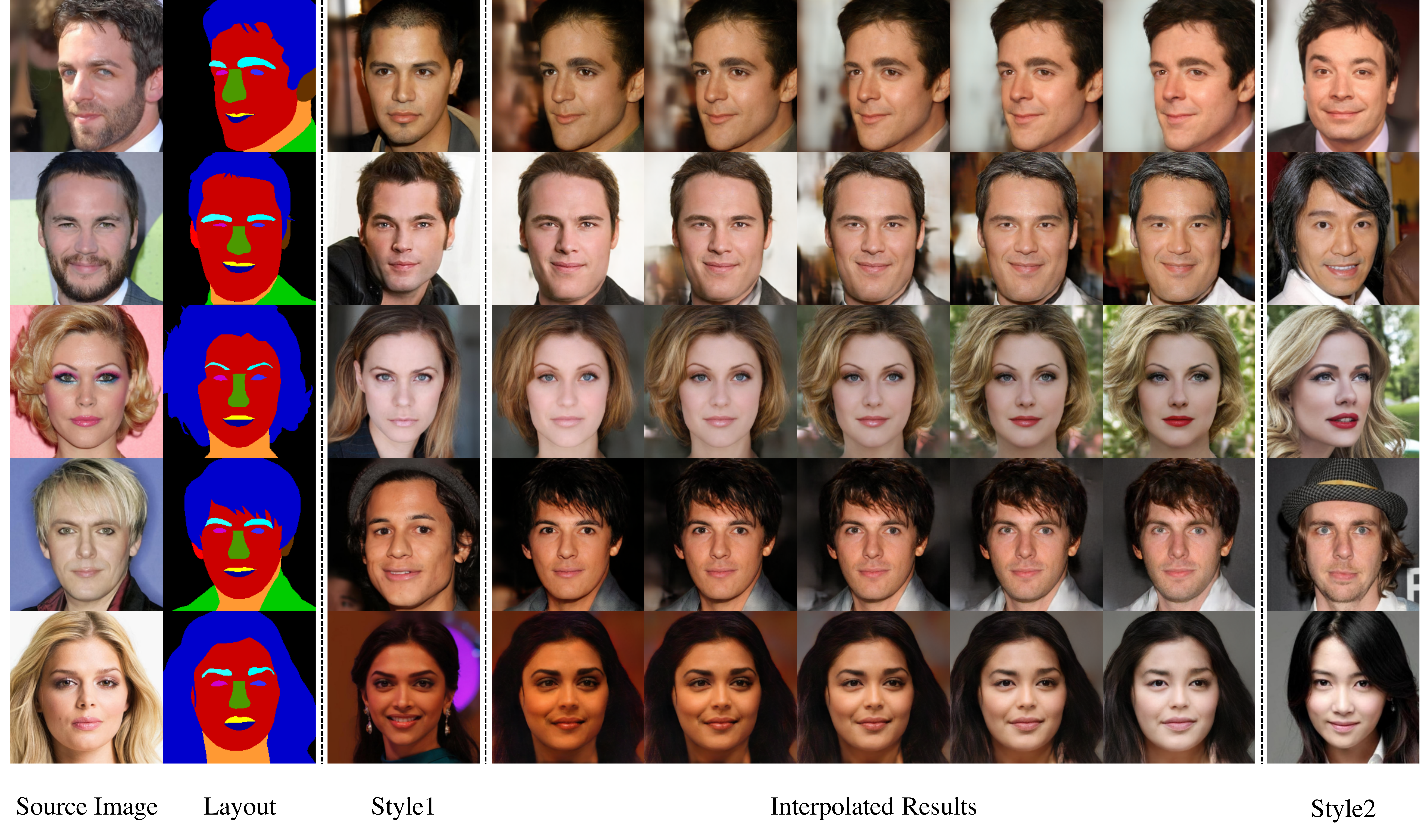}
\caption{Style interpolation on CelebAMask-HQ dataset}
\label{fig:Supp face interpolation}
\end{figure*}

\begin{figure*}[th]
\centering
\includegraphics[width=\linewidth]{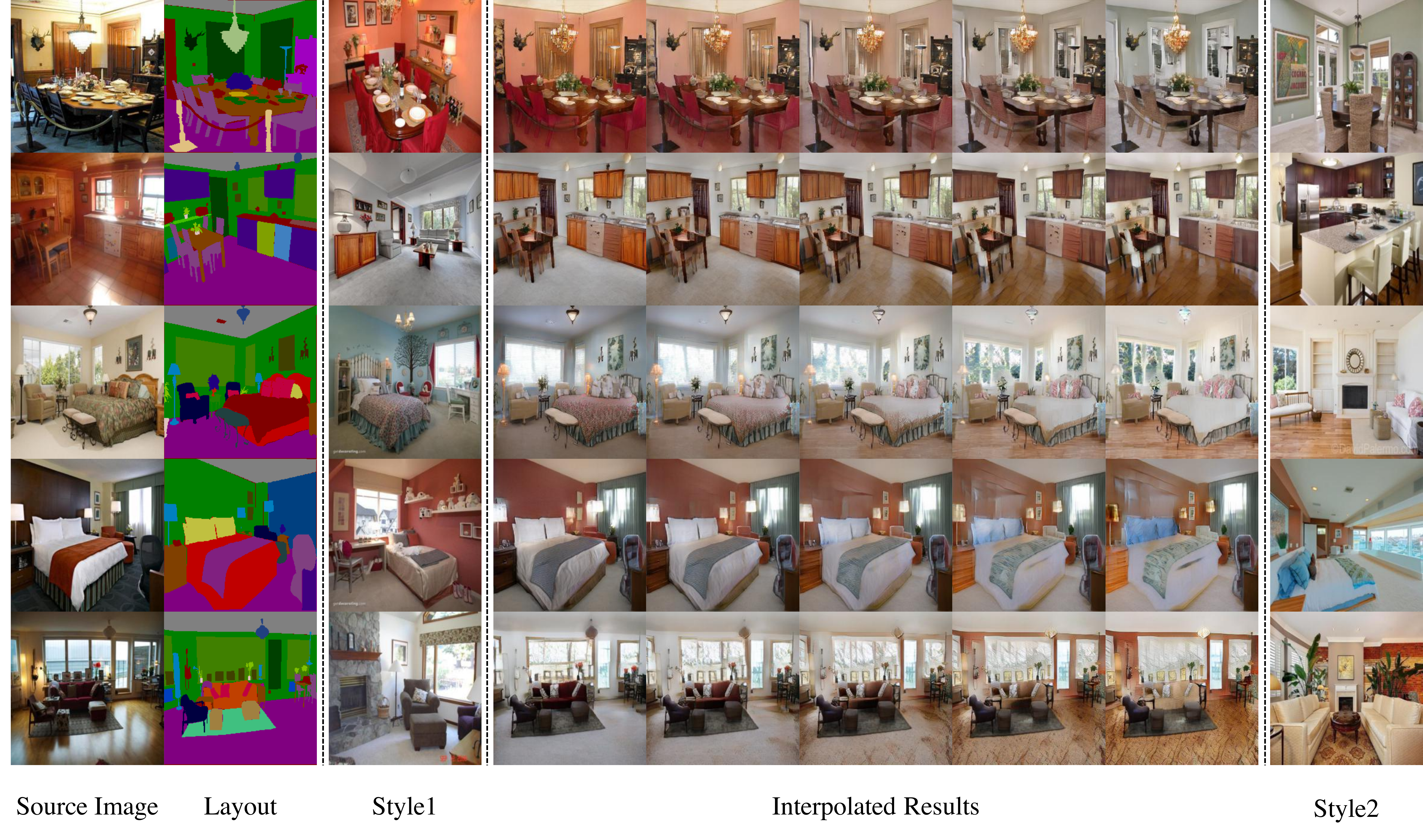}
\caption{Style interpolation on ADE20K dataset}
\label{fig:Supp ADE interpolation}
\end{figure*}

\addtolength{\tabcolsep}{-4.5pt}    
\bgroup
\def\arraystretch{0.5}
\begin{figure*}[]
\begin{tabular} {cc|cc|c}
 Label & Ground Truth & Pix2PixHD~\cite{wang2018pix2pixHD} &  SPADE~\cite{park2019SPADE} & Ours\\
\includegraphics[width=0.1932\textwidth]{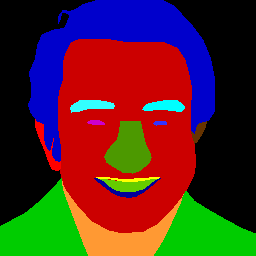} & \includegraphics[width=0.1932\textwidth]{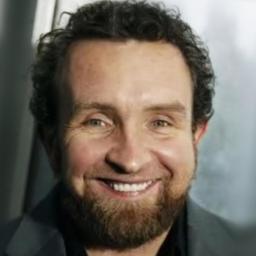} &
\includegraphics[width=0.1932\textwidth]{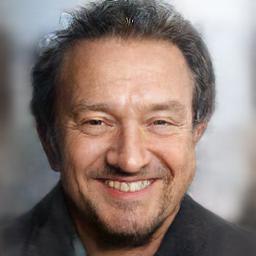} &   \includegraphics[width=0.1932\textwidth]{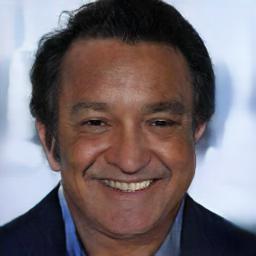} &  \includegraphics[width=0.1932\textwidth]{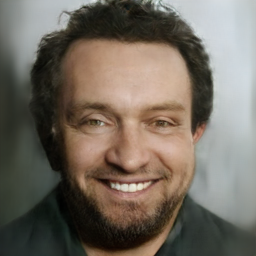} \\

\includegraphics[width=0.1932\textwidth]{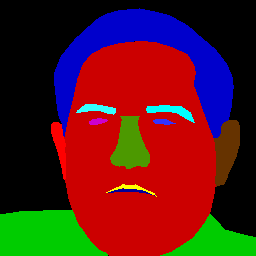} & \includegraphics[width=0.1932\textwidth]{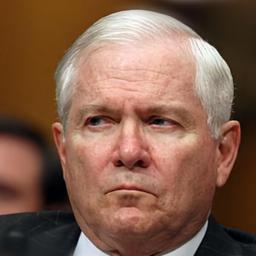} &
\includegraphics[width=0.1932\textwidth]{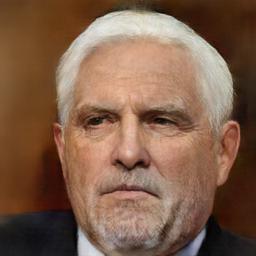}&
\includegraphics[width=0.1932\textwidth]{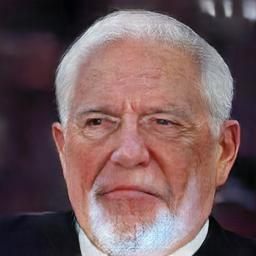} &  \includegraphics[width=0.1932\textwidth]{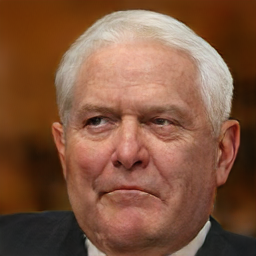} \\

\includegraphics[width=0.1932\textwidth]{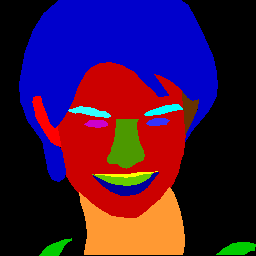} & \includegraphics[width=0.1932\textwidth]{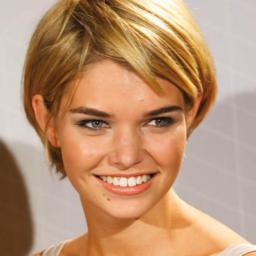} &
\includegraphics[width=0.1932\textwidth]{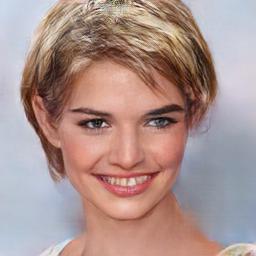}&
\includegraphics[width=0.1932\textwidth]{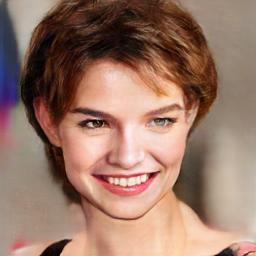} &  \includegraphics[width=0.1932\textwidth]{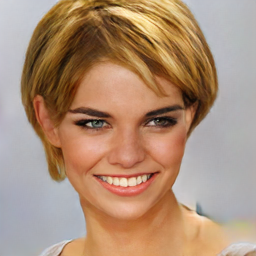} \\

\includegraphics[width=0.1932\textwidth]{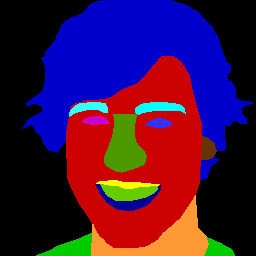} & \includegraphics[width=0.1932\textwidth]{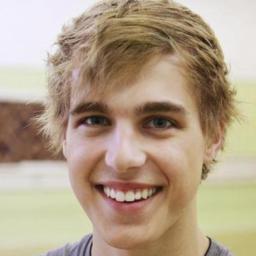} &
\includegraphics[width=0.1932\textwidth]{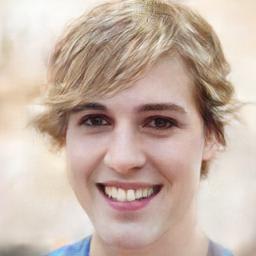} &   \includegraphics[width=0.1932\textwidth]{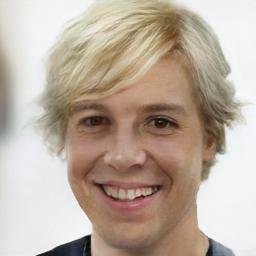} &  \includegraphics[width=0.1932\textwidth]{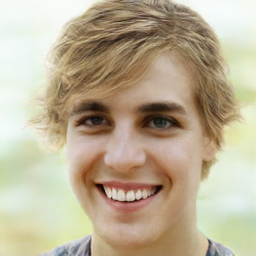} \\

\includegraphics[width=0.1932\textwidth]{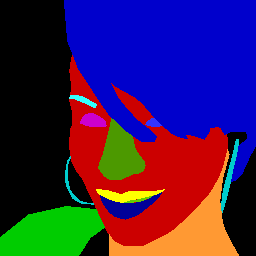} & \includegraphics[width=0.1932\textwidth]{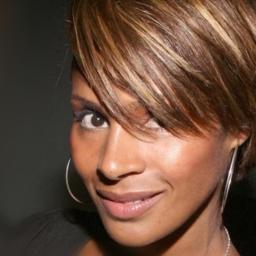} &
\includegraphics[width=0.1932\textwidth]{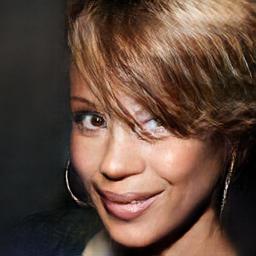} &   \includegraphics[width=0.1932\textwidth]{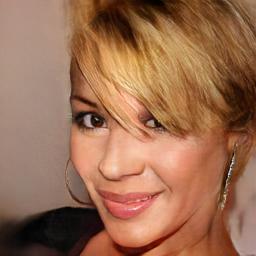} &  \includegraphics[width=0.1932\textwidth]{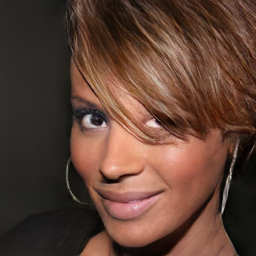} \\

\includegraphics[width=0.1932\textwidth]{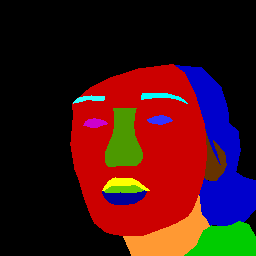} & \includegraphics[width=0.1932\textwidth]{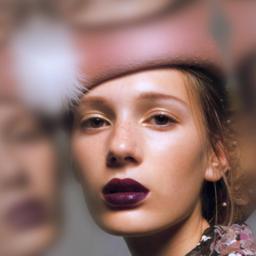} &
\includegraphics[width=0.1932\textwidth]{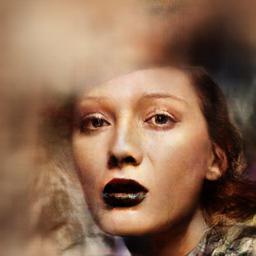} &   \includegraphics[width=0.1932\textwidth]{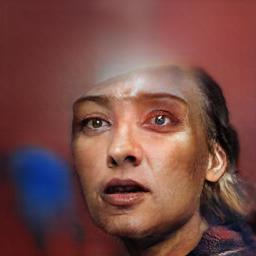} &  \includegraphics[width=0.1932\textwidth]{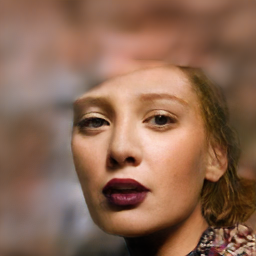} \\

\end{tabular}
\vspace{-2mm}
	\caption{Visual  comparison  of  semantic  image  synthesis  results  on  the  CelebAMask-HQ dataset. We compare Pix2PixHD, SPADE, and our method.}
	\label{fig:CelebAMask-HQ results}	
\vspace{-3mm}	
 \end{figure*}
 \egroup
 \addtolength{\tabcolsep}{4.5pt}

\addtolength{\tabcolsep}{-4.5pt}    
\bgroup
\def\arraystretch{0.5}
\begin{figure*}[]
\begin{tabular} {cc|cc|c}
 Label & Ground Truth & Pix2PixHD~\cite{wang2018pix2pixHD} &  SPADE~\cite{park2019SPADE} & Ours\\

\includegraphics[width=0.1932\textwidth,height=0.96in]{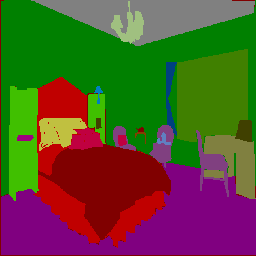} & \includegraphics[width=0.1932\textwidth,height=0.96in]{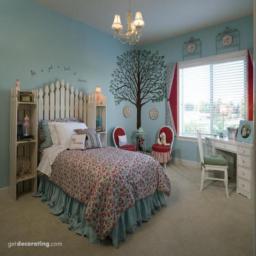} &
\includegraphics[width=0.1932\textwidth,height=0.96in]{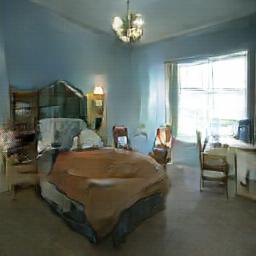} &   \includegraphics[width=0.1932\textwidth,height=0.96in]{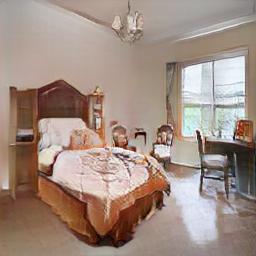} &  \includegraphics[width=0.1932\textwidth,height=0.96in]{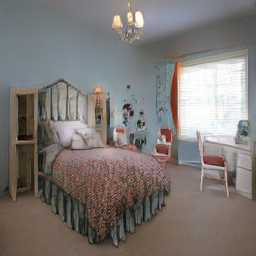} \\

\includegraphics[width=0.1932\textwidth,height=0.96in]{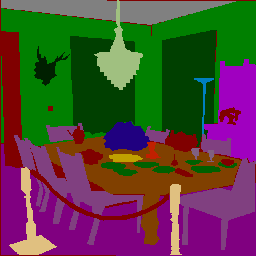} & \includegraphics[width=0.1932\textwidth,height=0.96in]{Images/Rec/ADE/gt/ADE_val_00000309.jpg} &
\includegraphics[width=0.1932\textwidth,height=0.96in]{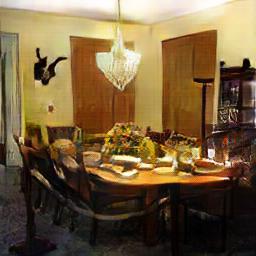} &   \includegraphics[width=0.1932\textwidth,height=0.96in]{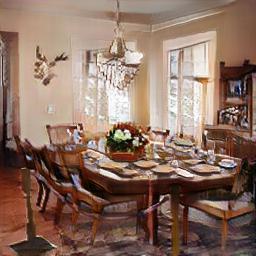} &  \includegraphics[width=0.1932\textwidth,height=0.96in]{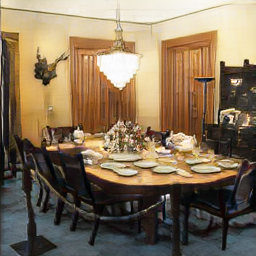} \\

\includegraphics[width=0.1932\textwidth,height=0.96in]{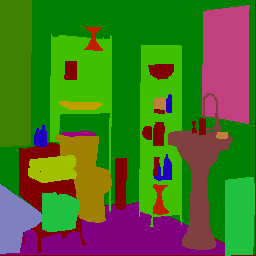} & \includegraphics[width=0.1932\textwidth,height=0.96in]{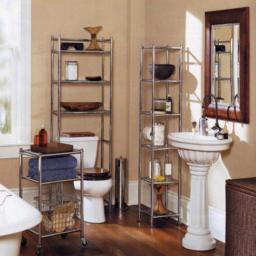} &
\includegraphics[width=0.1932\textwidth,height=0.96in]{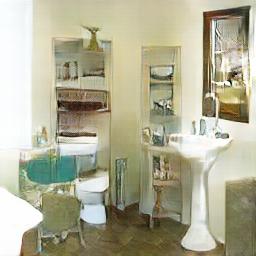} &   \includegraphics[width=0.1932\textwidth,height=0.96in]{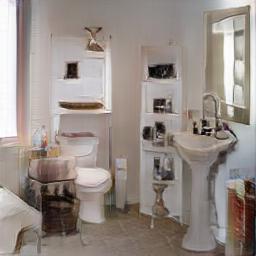} &  \includegraphics[width=0.1932\textwidth,height=0.96in]{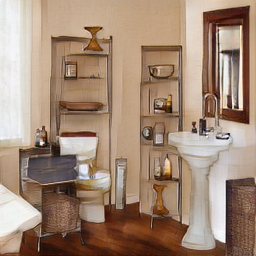} \\

\includegraphics[width=0.1932\textwidth,height=0.96in]{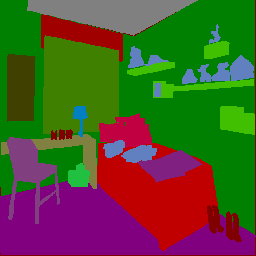} & \includegraphics[width=0.1932\textwidth,height=0.96in]{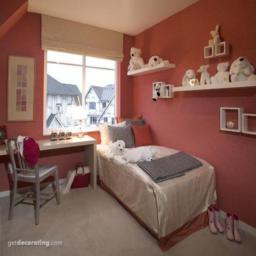} &
\includegraphics[width=0.1932\textwidth,height=0.96in]{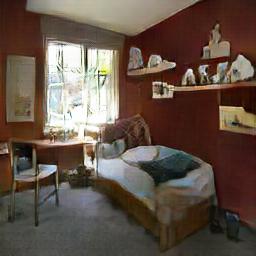} &   \includegraphics[width=0.1932\textwidth,height=0.96in]{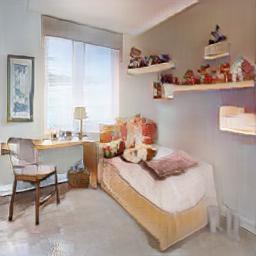} &  \includegraphics[width=0.1932\textwidth,height=0.96in]{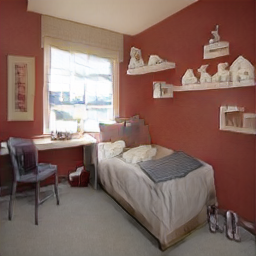} \\

\includegraphics[width=0.1932\textwidth,height=0.96in]{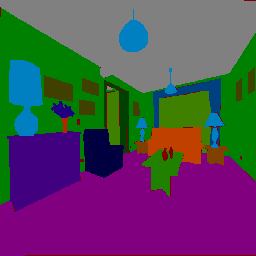} & \includegraphics[width=0.1932\textwidth,height=0.96in]{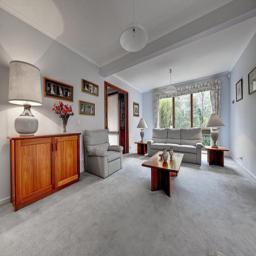} &
\includegraphics[width=0.1932\textwidth,height=0.96in]{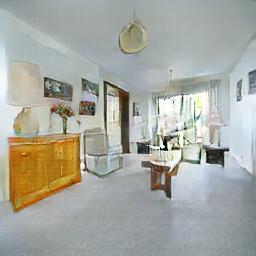} &   \includegraphics[width=0.1932\textwidth,height=0.96in]{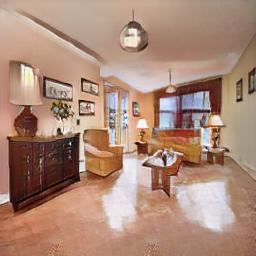} &  \includegraphics[width=0.1932\textwidth,height=0.96in]{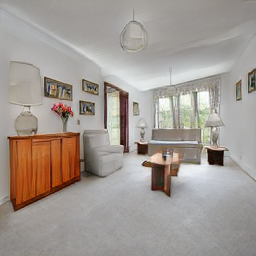} \\

\includegraphics[width=0.1932\textwidth,height=0.96in]{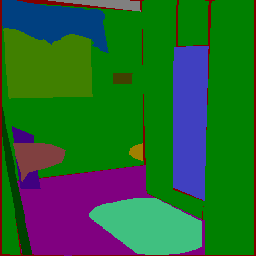} & \includegraphics[width=0.1932\textwidth,height=0.96in]{Images/Rec/ADE/gt/ADE_val_00000090.jpg} &
\includegraphics[width=0.1932\textwidth,height=0.96in]{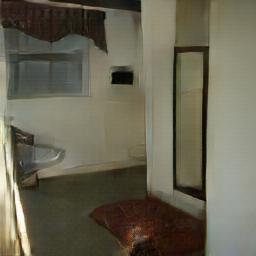} &   \includegraphics[width=0.1932\textwidth,height=0.96in]{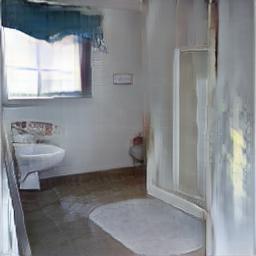} &  \includegraphics[width=0.1932\textwidth,height=0.96in]{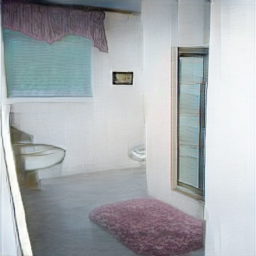} \\

\includegraphics[width=0.1932\textwidth,height=0.96in]{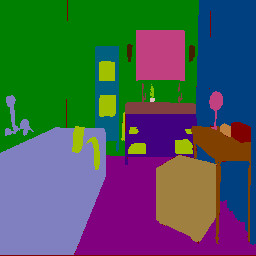} & \includegraphics[width=0.1932\textwidth,height=0.96in]{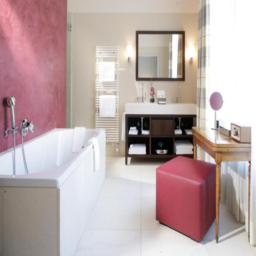} &
\includegraphics[width=0.1932\textwidth,height=0.96in]{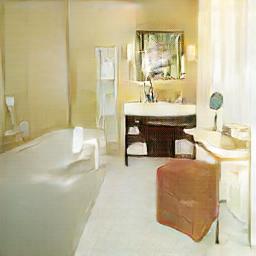} &   \includegraphics[width=0.1932\textwidth,height=0.96in]{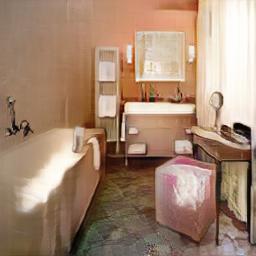} &  \includegraphics[width=0.1932\textwidth,height=0.96in]{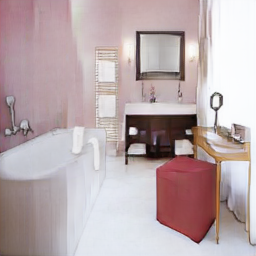} \\

\includegraphics[width=0.1932\textwidth,height=0.96in]{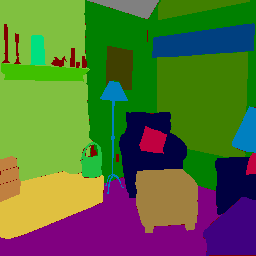} & \includegraphics[width=0.1932\textwidth,height=0.96in]{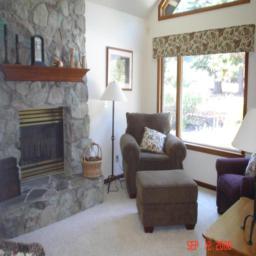} &
\includegraphics[width=0.1932\textwidth,height=0.96in]{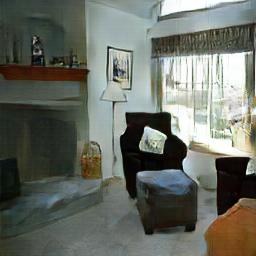} &   \includegraphics[width=0.1932\textwidth,height=0.96in]{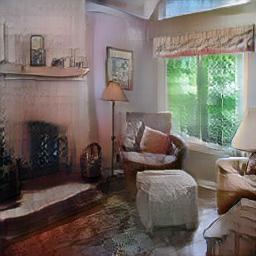} &  \includegraphics[width=0.1932\textwidth,height=0.96in]{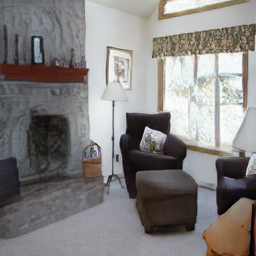} \\

\end{tabular}
\vspace{-2mm}
	\caption{Visual  comparison  of  semantic  image  synthesis  results  on  the ADE20K dataset. We compare Pix2PixHD, SPADE, and our method.}
	\label{fig:ADE20K results}	
\vspace{-3mm}	
 \end{figure*}
 \egroup
 \addtolength{\tabcolsep}{4.5pt}

\addtolength{\tabcolsep}{-4.5pt}    
\bgroup
\def\arraystretch{0.5}
\begin{figure*}[]
\begin{tabular} {cc|cc|c}
 Label & Ground Truth & Pix2PixHD~\cite{wang2018pix2pixHD} &  SPADE~\cite{park2019SPADE} & Ours\\

\includegraphics[width=0.1932\textwidth,height=0.96in]{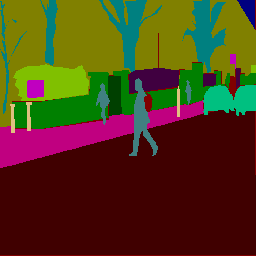} &
\includegraphics[width=0.1932\textwidth,height=0.96in]{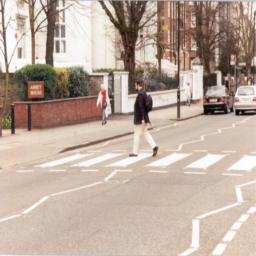} & 
\includegraphics[width=0.1932\textwidth,height=0.96in]{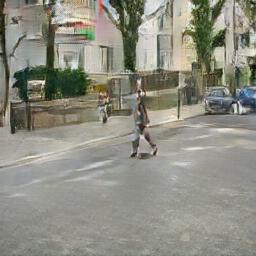} &   \includegraphics[width=0.1932\textwidth,height=0.96in]{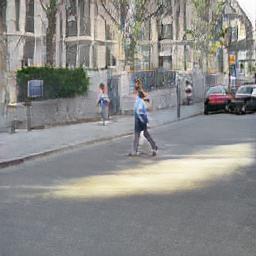} &  \includegraphics[width=0.1932\textwidth,height=0.96in]{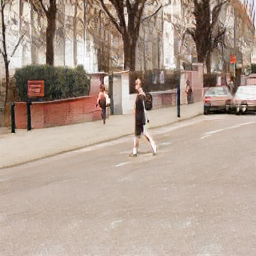} \\

\includegraphics[width=0.1932\textwidth,height=0.96in]{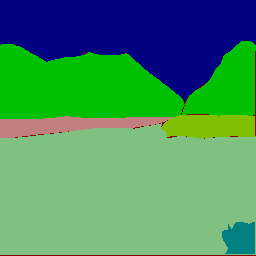} &
\includegraphics[width=0.1932\textwidth,height=0.96in]{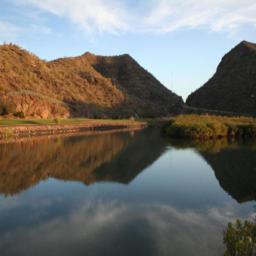} & 
\includegraphics[width=0.1932\textwidth,height=0.96in]{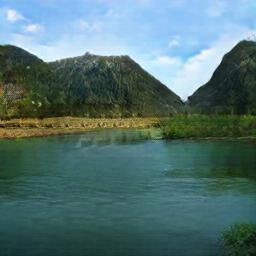} &   \includegraphics[width=0.1932\textwidth,height=0.96in]{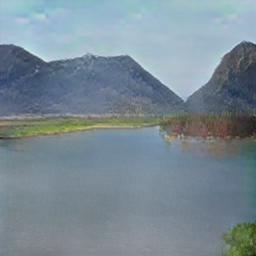} &  \includegraphics[width=0.1932\textwidth,height=0.96in]{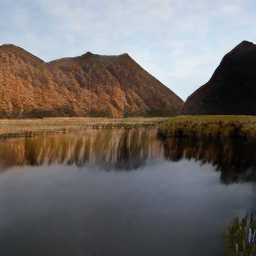} \\

\includegraphics[width=0.1932\textwidth,height=0.96in]{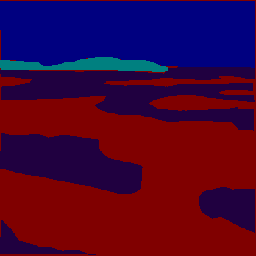} &
\includegraphics[width=0.1932\textwidth,height=0.96in]{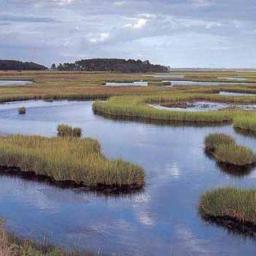} & 
\includegraphics[width=0.1932\textwidth,height=0.96in]{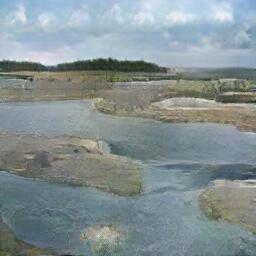} &   \includegraphics[width=0.1932\textwidth,height=0.96in]{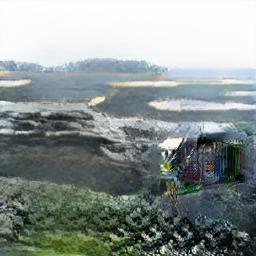} &  \includegraphics[width=0.1932\textwidth,height=0.96in]{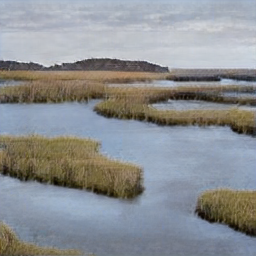} \\

\includegraphics[width=0.1932\textwidth,height=0.96in]{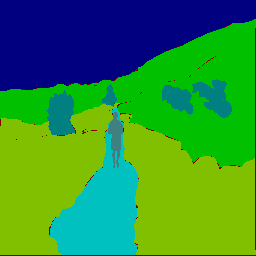} & \includegraphics[width=0.1932\textwidth,height=0.96in]{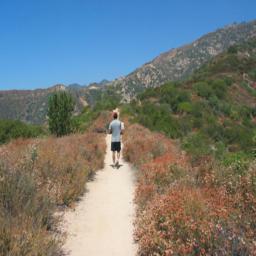} &
\includegraphics[width=0.1932\textwidth,height=0.96in]{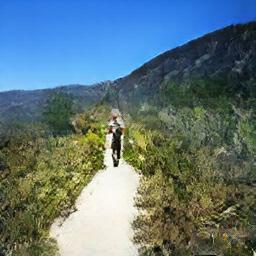} &   \includegraphics[width=0.1932\textwidth,height=0.96in]{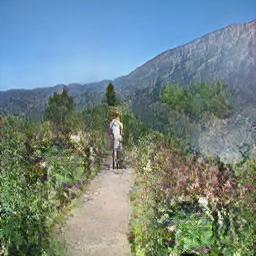} &  \includegraphics[width=0.1932\textwidth,height=0.96in]{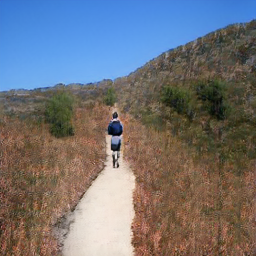} \\

\includegraphics[width=0.1932\textwidth,height=0.96in]{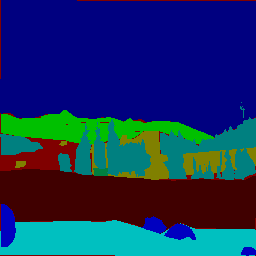} & \includegraphics[width=0.1932\textwidth,height=0.96in]{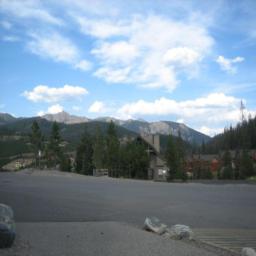} &
\includegraphics[width=0.1932\textwidth,height=0.96in]{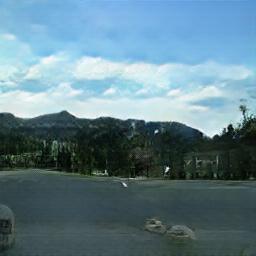} &   \includegraphics[width=0.1932\textwidth,height=0.96in]{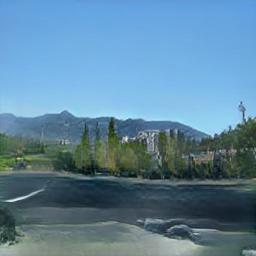} &  \includegraphics[width=0.1932\textwidth,height=0.96in]{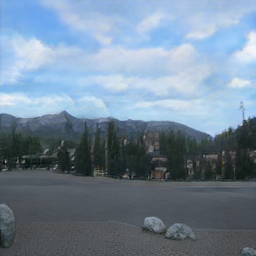} \\

\includegraphics[width=0.1932\textwidth,height=0.96in]{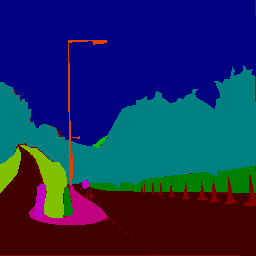} & \includegraphics[width=0.1932\textwidth,height=0.96in]{Images/Rec/ADE_outdoor/gt/ADE_val_00000662.jpg} &
\includegraphics[width=0.1932\textwidth,height=0.96in]{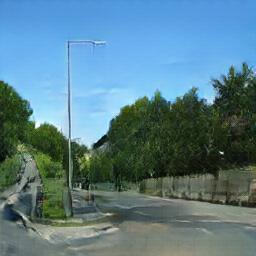} &   \includegraphics[width=0.1932\textwidth,height=0.96in]{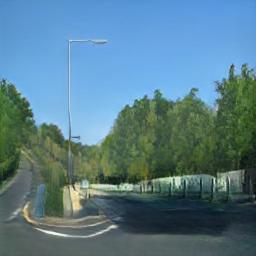} &  \includegraphics[width=0.1932\textwidth,height=0.96in]{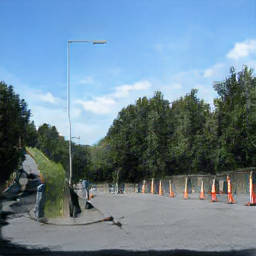} \\

\includegraphics[width=0.1932\textwidth,height=0.96in]{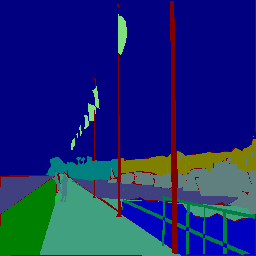} & \includegraphics[width=0.1932\textwidth,height=0.96in]{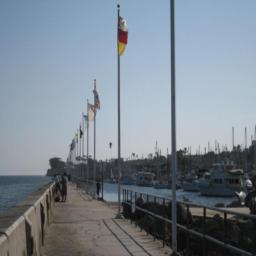} &
\includegraphics[width=0.1932\textwidth,height=0.96in]{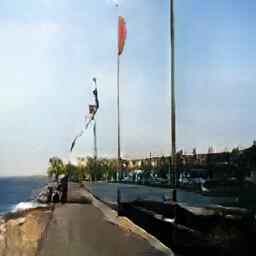} &   \includegraphics[width=0.1932\textwidth,height=0.96in]{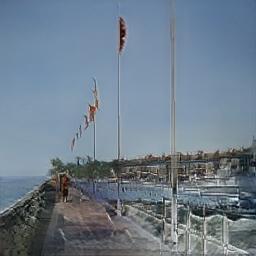} &  \includegraphics[width=0.1932\textwidth,height=0.96in]{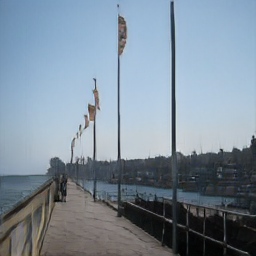} \\

\includegraphics[width=0.1932\textwidth,height=0.96in]{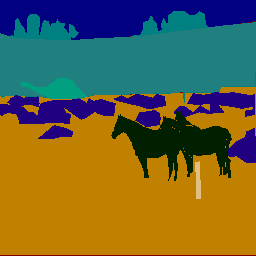} & \includegraphics[width=0.1932\textwidth,height=0.96in]{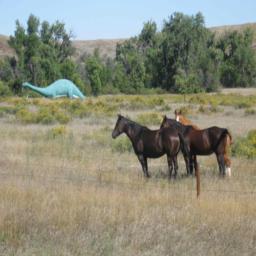} &
\includegraphics[width=0.1932\textwidth,height=0.96in]{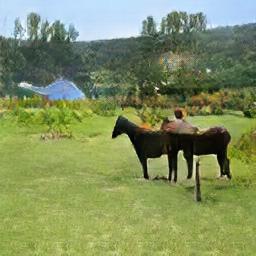} &   \includegraphics[width=0.1932\textwidth,height=0.96in]{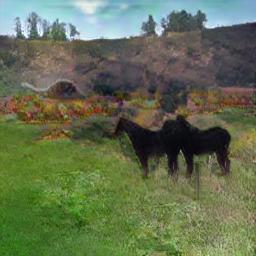} &  \includegraphics[width=0.1932\textwidth,height=0.96in]{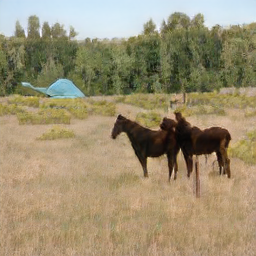} \\

\end{tabular}
\vspace{-2mm}
	\caption{Visual  comparison  of  semantic  image  synthesis  results  on  the ADE20K dataset. We compare Pix2PixHD, SPADE, and our method.}
	\label{fig:ADE20K outdoor results}	
\vspace{-3mm}	
 \end{figure*}
 \egroup
 \addtolength{\tabcolsep}{4.5pt}

\addtolength{\tabcolsep}{-4.5pt}    
\bgroup
\def\arraystretch{0.5}
\begin{figure*}[]
\begin{tabular} {cc|cc|c}
 Label & Ground Truth & Pix2PixHD~\cite{wang2018pix2pixHD} &  SPADE~\cite{park2019SPADE} & Ours\\

\includegraphics[width=0.1932\textwidth]{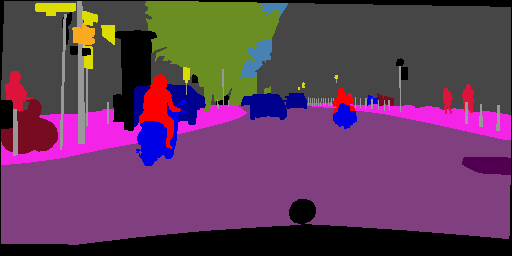} & \includegraphics[width=0.1932\textwidth]{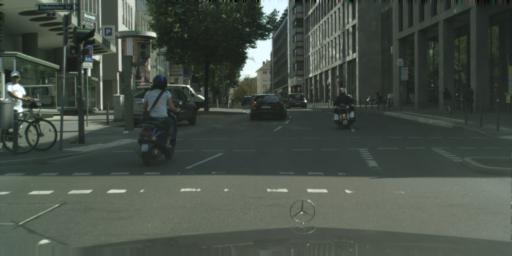} &
\includegraphics[width=0.1932\textwidth]{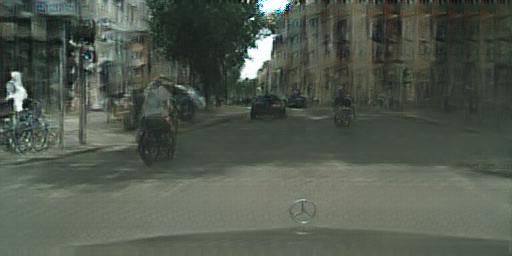} &   \includegraphics[width=0.1932\textwidth]{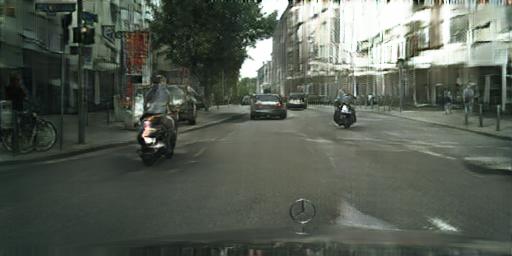} &  \includegraphics[width=0.1932\textwidth]{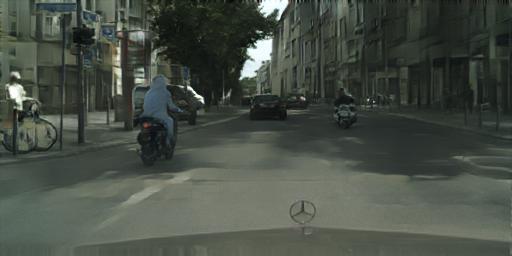} \\

\includegraphics[width=0.1932\textwidth]{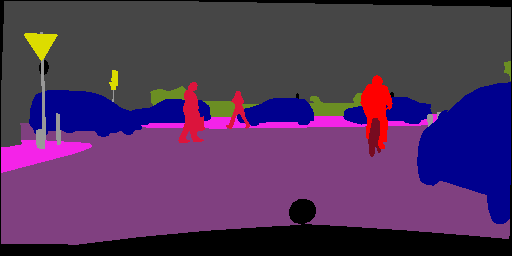} & \includegraphics[width=0.1932\textwidth]{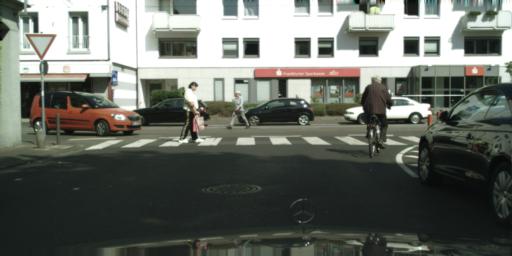} &
\includegraphics[width=0.1932\textwidth]{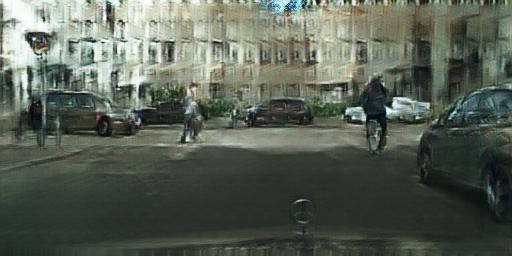} &   \includegraphics[width=0.1932\textwidth]{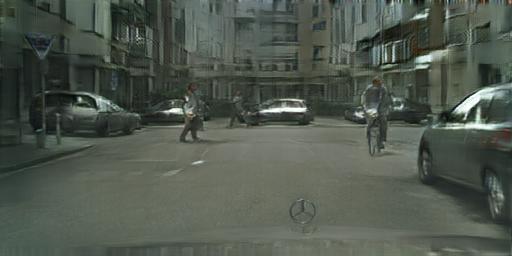} &  \includegraphics[width=0.1932\textwidth]{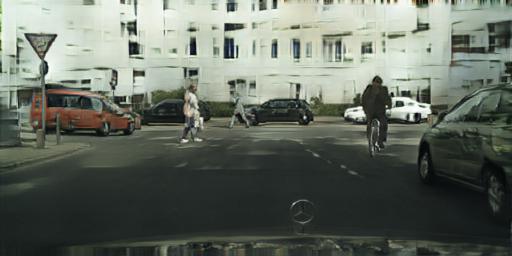} \\

\includegraphics[width=0.1932\textwidth]{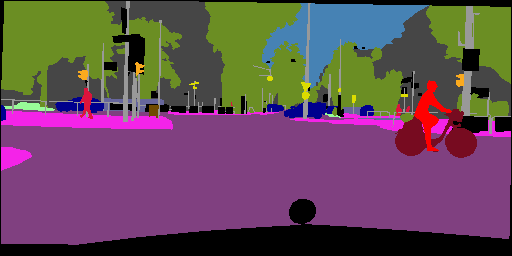} & \includegraphics[width=0.1932\textwidth]{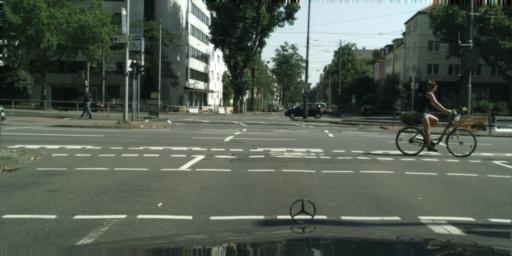} &
\includegraphics[width=0.1932\textwidth]{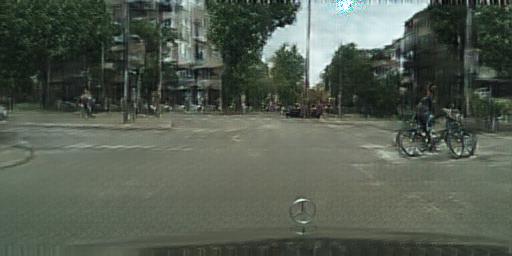} &   \includegraphics[width=0.1932\textwidth]{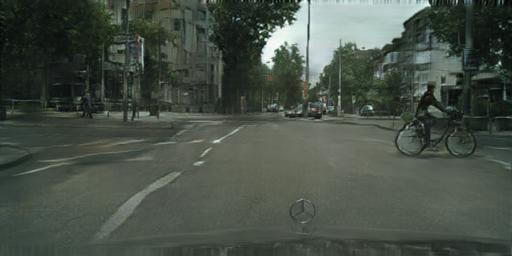} &  \includegraphics[width=0.1932\textwidth]{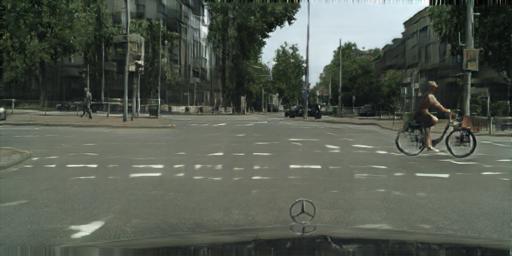} \\

\includegraphics[width=0.1932\textwidth]{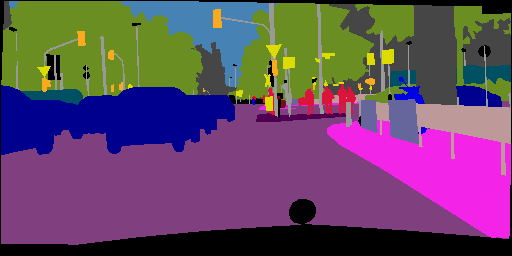} & \includegraphics[width=0.1932\textwidth]{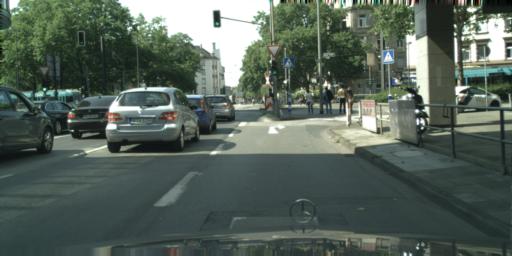} &
\includegraphics[width=0.1932\textwidth]{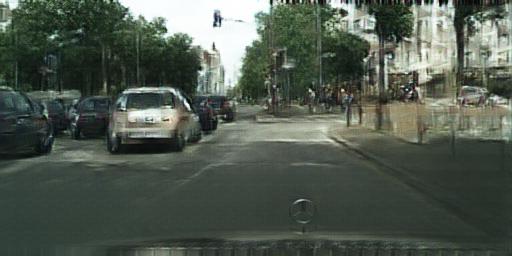} &   \includegraphics[width=0.1932\textwidth]{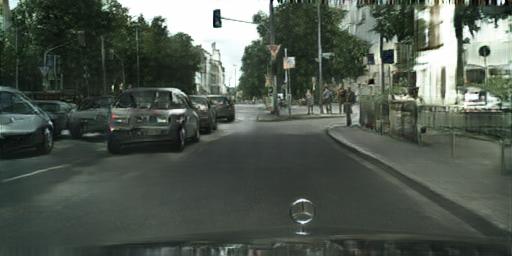} &  \includegraphics[width=0.1932\textwidth]{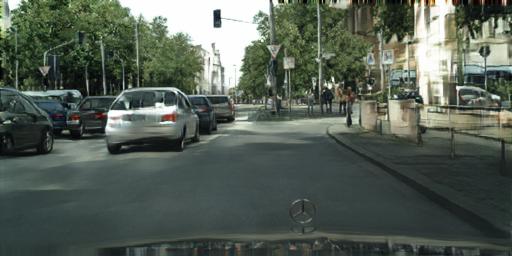} \\

\includegraphics[width=0.1932\textwidth]{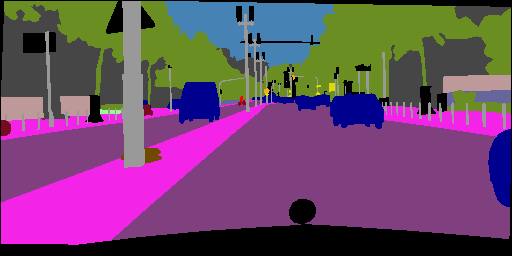} & \includegraphics[width=0.1932\textwidth]{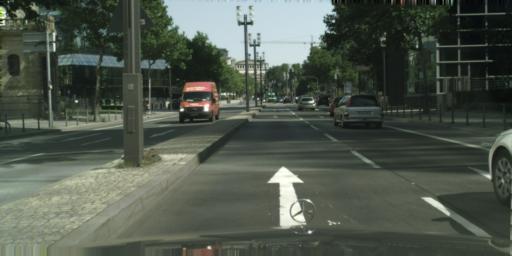} &
\includegraphics[width=0.1932\textwidth]{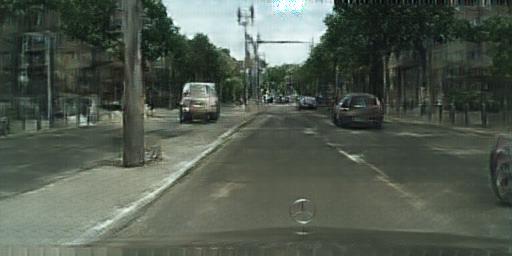} &   \includegraphics[width=0.1932\textwidth]{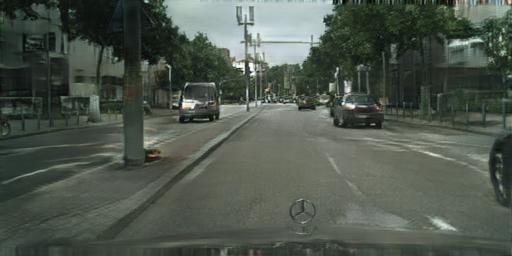} &  \includegraphics[width=0.1932\textwidth]{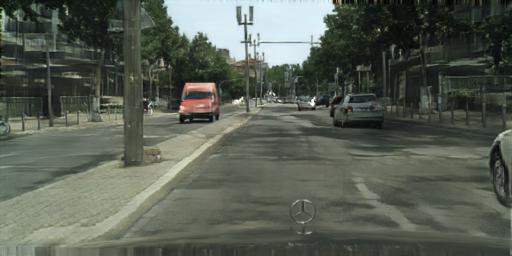} \\

\includegraphics[width=0.1932\textwidth]{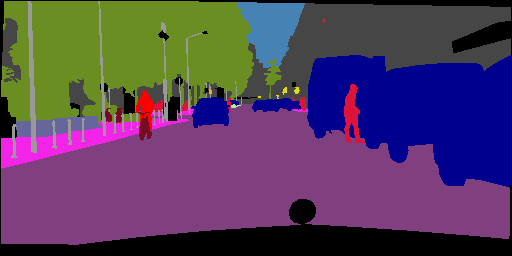} & \includegraphics[width=0.1932\textwidth]{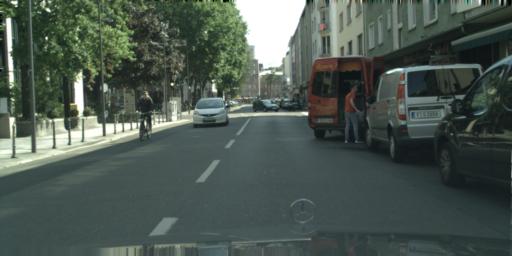} &
\includegraphics[width=0.1932\textwidth]{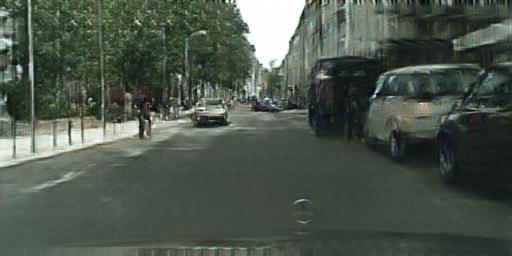} &   \includegraphics[width=0.1932\textwidth]{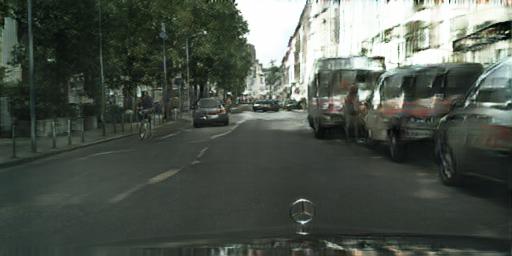} &  \includegraphics[width=0.1932\textwidth]{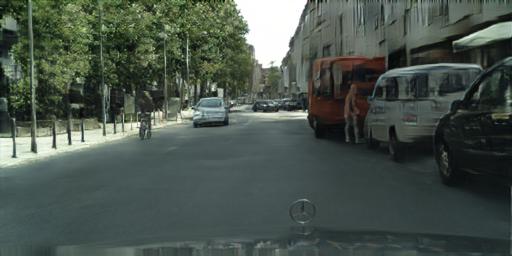} \\

\includegraphics[width=0.1932\textwidth]{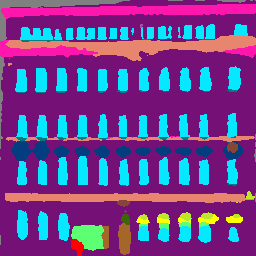} & \includegraphics[width=0.1932\textwidth]{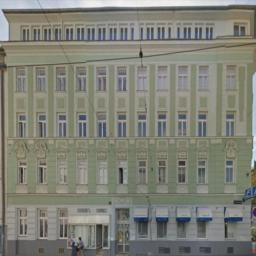} &
\includegraphics[width=0.1932\textwidth]{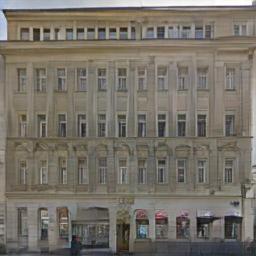} &   \includegraphics[width=0.1932\textwidth]{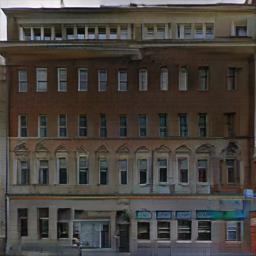} &  \includegraphics[width=0.1932\textwidth]{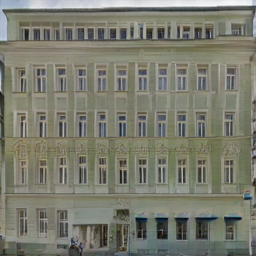} \\

\includegraphics[width=0.1932\textwidth]{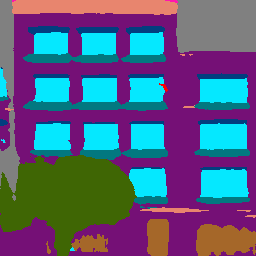} & 
\includegraphics[width=0.1932\textwidth]{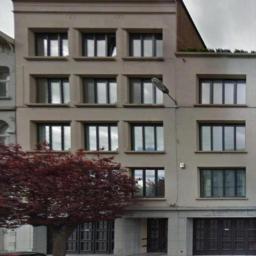} &
\includegraphics[width=0.1932\textwidth]{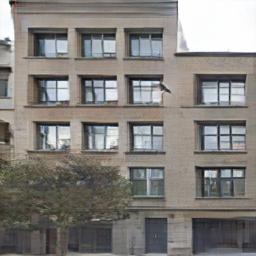} & 
\includegraphics[width=0.1932\textwidth]{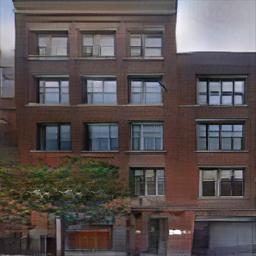} & 
\includegraphics[width=0.1932\textwidth]{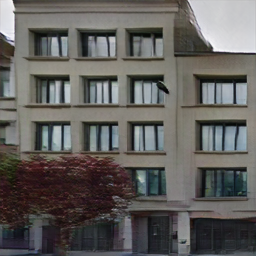} \\

\includegraphics[width=0.1932\textwidth]{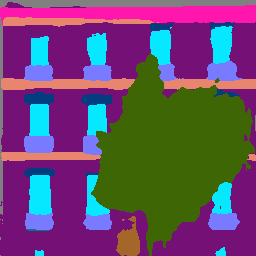} & 
\includegraphics[width=0.1932\textwidth]{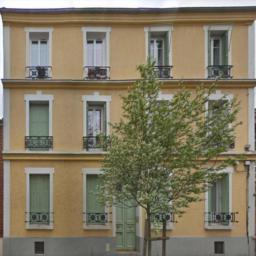} &
\includegraphics[width=0.1932\textwidth]{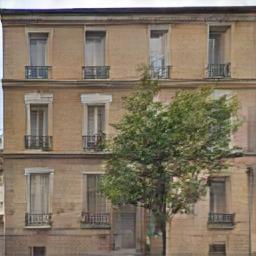} & 
\includegraphics[width=0.1932\textwidth]{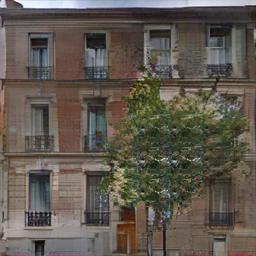} & 
\includegraphics[width=0.1932\textwidth]{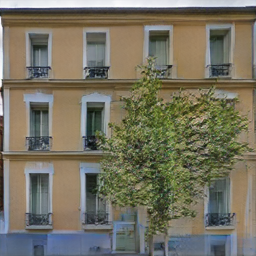} \\

\end{tabular}
\vspace{-2mm}
	\caption{Visual  comparison  of  semantic  image  synthesis  results  on  the  CityScapes and \Facades dataset. We compare Pix2PixHD, SPADE, and our method.}
	\label{fig:City Facades results}	
\vspace{-3mm}	
 \end{figure*}
 \egroup
 \addtolength{\tabcolsep}{4.5pt}




\end{document}